\documentclass[twocolumn]{article}
\usepackage[utf8]{inputenc}

\usepackage{graphicx}

\usepackage{colortbl}
\usepackage[numbers]{natbib}
\usepackage{comment}
\usepackage{balance}
\usepackage{enumitem}
\usepackage{makecell}
\usepackage{array}
\usepackage{dirtytalk}

\usepackage{subfig}
\usepackage{amssymb}
\usepackage{amsmath}

\usepackage[final]{microtype}

\usepackage{algorithm}
\usepackage{algpseudocode}
\usepackage{fancyhdr}
\usepackage{ragged2e}

\newcolumntype{L}[1]{>{\raggedright\let\newline\\\arraybackslash\hspace{0pt}}m{#1}}
\newcolumntype{C}[1]{>{\centering\let\newline\\\arraybackslash\hspace{0pt}}m{#1}}
\newcolumntype{R}[1]{>{\raggedleft\let\newline\\\arraybackslash\hspace{0pt}}m{#1}}

\fancypagestyle{firstpage}
{
    \fancyhf{}

    \fancyhead[L]{\small \justifying \noindent \textbf{Cite as: Bignold, A., Cruz, F., Dazeley, R., Vamplew, P., Foale, C. (2021). Persistent Rule-based Interactive Reinforcement Learning. Neural Computing and Applications.} 
    }

    \fancyfoot[L]{\small \justifying \noindent This preprint has not undergone peer review or any post-submission improvements or corrections. The Version of Record of this article is published in Neural Computing and Applications, and is available online at https://doi.org/10.1007/s00521-021-06466-w.
    }
    
}

\title{Persistent Rule-based Interactive Reinforcement Learning}
\author{
Adam Bignold$^{1,*}$ \and
Francisco Cruz$^{2,3,*}$ \and
Richard Dazeley$^{2}$ \and 
Peter Vamplew$^{1}$ \and
Cameron Foale$^{1}$
}
\date{\normalsize
$^1$ School of Science, Engineering and Information Technology, Federation University, Ballarat, Australia.\\
$^2$ School of Information Technology, Deakin University, Geelong, Australia.\\
$^3$ Escuela de Ingenier\'ia, Universidad Central de Chile, Santiago, Chile.\\
$^*$ Both authors contributed equally to this manuscript.\\
Corresponding e-mails: \{a.bignold, p.vamplew, c.foale\}@federation.edu.au, \\ \{francisco.cruz, richard.dazeley\}@deakin.edu.au\\
}

\begin{document}

\maketitle

\begin{abstract}
Interactive reinforcement learning has allowed speeding up the learning process in autonomous agents by including a human trainer providing extra information to the agent in real-time. 
Current interactive reinforcement learning research has been limited to real-time interactions that offer relevant user advice to the current state only. 
Additionally, the information provided by each interaction is not retained and instead discarded by the agent after a single-use. 
In this work, we propose a persistent rule-based interactive reinforcement learning approach, i.e., a method for retaining and reusing provided knowledge, allowing trainers to give general advice relevant to more than just the current state. 
Our experimental results show persistent advice substantially improves the performance of the agent while reducing the number of interactions required for the trainer. 
Moreover, rule-based advice shows similar performance impact as state-based advice, but with a substantially reduced interaction count.
\end{abstract}


\thispagestyle{firstpage}

\section{Introduction}

Interactive reinforcement learning (IntRL) allows a trainer to guide or evaluate a learning agent's behaviour~\cite{arzate2020survey, lin2020review}. 
The assistance provided by the trainer reinforces the behaviour the agent is learning and shapes the exploration policy, resulting in a reduced search space~\cite{bignold2020human}. 
Current IntRL techniques discard the advice sourced from the human shortly after it has been used~\cite{knox2009interactively, bignold2020conceptual}, increasing the dependency on the advisor to repeatedly provide the same advice to maximise the agent’s use of it.

\begin{figure*}
\centering
\includegraphics[width=0.7\linewidth]{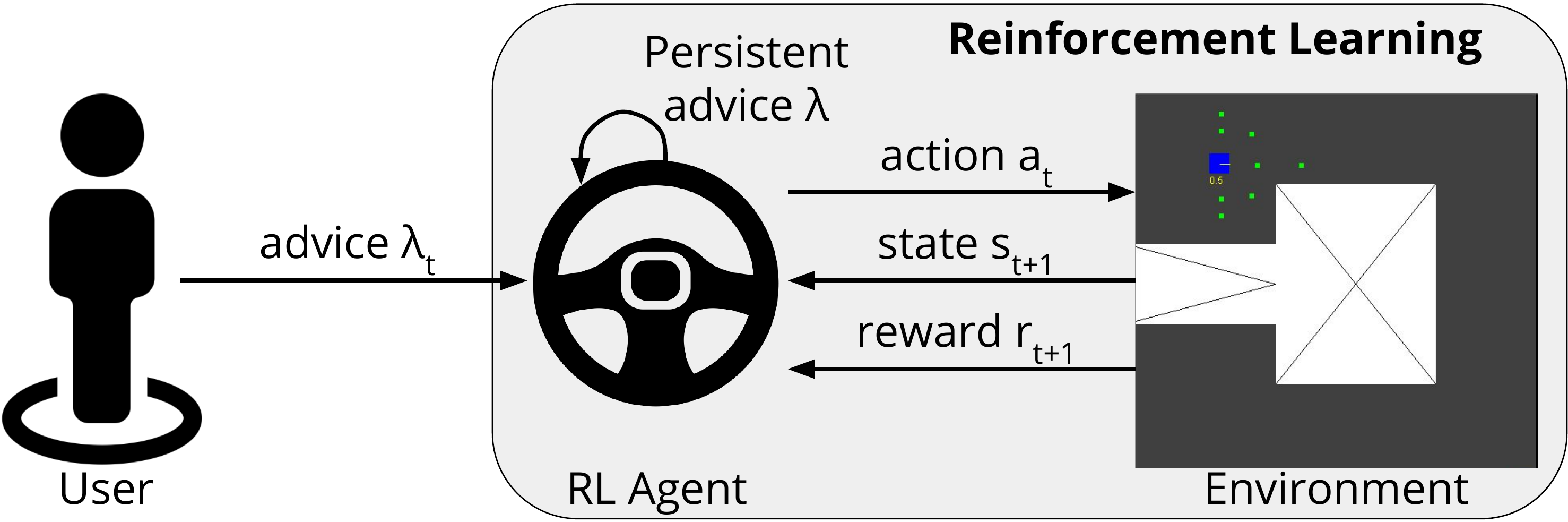}
\caption{Interactive reinforcement learning framework.
In traditional RL an agent performs an action and observes a new state and reward.
In the figure, the environment is represented by the simulated self-driving car scenario and the RL agent may control the direction and speed of the car.
IntRL adds advice from a user acting as an external expert in certain situations.
Our proposal includes the use of persistent rule-based advice in order to minimise the interaction with the trainer.
}
\label{fig:irl}
\end{figure*}

Moreover, current IntRL approaches allow trainers to evaluate or recommend actions based only on the current state of the environment~\cite{policyshaping, tamer}. 
This constraint restricts the trainer to providing advice relevant to the current state and no other, even when such advice may be applicable to multiple states~\cite{taylor2014reinforcement}. 
Restricting the time and utility of advice affect negatively the interactive approach in terms of creating an increasing demand on the user's time, instead of withholding potentially useful information for the agent~\cite{lin2020review}.
In this regard, interaction is advice received from a trainer agent, and this trainer may be either a human, a simulated user, an intelligent agent previously trained, or an oracle with full knowledge of the intended task~\cite{bignold2020conceptual}.

This work introduces persistence to IntRL, a method for information retention and reuse. 
Persistent agents attempt to maximise the value extracted from the advice by replaying interactions that occurred in the past, rather than relying on the advisor to repeat an interaction. 
Agents that retain advice require fewer interactions than non-persistent counterparts to achieve similar or improved performance, thus reducing the burden on the advisor to provide advice on an ongoing basis. 

Additionally, this works introduces a persistent rule-based IntRL approach. 
Allowing users to provide information in the form of rules, rather than per-state action recommendations, increases the information per interaction, and does not limit the information to the current state. 
By not constraining the advice to the current state, users can give advice pre-emptively, no longer requiring the current state to match the criteria for the user's assistance. 
This more informationally rich interaction method improves the performance of the agent compared to existing methods and reduces the number of interactions between the agent and the advisor. 
Considering that advice in IntRL comes from expert users, the total amount of interaction is relevant. A lower dependency on experts makes the approach feasible to work in real-world scenarios (e.g., human-robot interaction) in which a human expert may be available for a very limited number of steps.

Therefore, the contribution of this work is twofold.
First, the introduction of a state-based method for the retention and reuse of advice, named persistence. 
Second, a persistent rule-based IntRL method obtaining the same performance as state-based advice, but with a substantially reduced interaction count. 
In this regard, rules allow advice to be provided that generalises over multiple states.

\section{Rule-based Interactive\\Advice}

\subsection{Reinforcement Learning and\\Interactive Advice}

Reinforcement learning (RL)~\cite{sutton2018reinforcement} is a machine learning technique that allows an agent to learn the dynamics of an environment by interacting with it.
When interacting the agent transits from the current state $s_t$ to a new state $s_{t+1}$ by performing action $a_t$.
Additionally, the agent receives a reward value $r_{t+1}$ for the actions performed.
During this process, the agent observes both the new states and the reward signal and learns a policy $\pi:S \rightarrow A$, where $S$ is the set of all possible states and $A$ the set of actions available from $S$.
Figure~\ref{fig:irl} shows the traditional RL loop between the agent and the environment inside the grey box.

An environment in an RL problem can be described as a Markov decision process (MDP)~\cite{Puterman94}.
An MDP in defined as a 4-tuple $<S, A, \delta, r>$, where $S$ is a finite set of states, $A$ is a set of actions, $\delta$ is the transition function $\delta:S \times A \rightarrow S$, and, $r$ is the reward function $r:S \times A \rightarrow \mathbb{R}$.

In an MDP, a state with the Markovian property contains all the information about the dynamics of the task, i.e., the next state and the reward depend only on the action selected.
Therefore, the history of previous transitions is not relevant in terms of the decision-making problem~\cite{sutton2018reinforcement}.
Thus, the probability that $s_t$, $r_t$, and $a_t$ take values $s'$, $r$ and $a$ with the previous state being $s$ is given by:

\begin{equation}
p(s',r|s,a) = P(s_t=s', r_t=r | s_{t-1} = s, a_{t-1}=a).
\end{equation}

By interacting with the environment, an agent has to deal with the exploration/exploitation trade-off problem, that is, the agent has to explore the action space offsetting the already explored good actions with others that it never tried~\cite{sledge2017balancing}.
Hence, the agent needs a strategy to choose actions to perform in a given state. 
An alternative is to use the $\epsilon$-greedy action selection method.
This method uses an exploration factor ($\epsilon$) which is randomly chosen from a uniform distribution.
The probability $P(s_t,a)$ of selection action $a$ in state $s_t$ can be formally defined as:

\begin{equation}
P(s_t,a) = \left\{
\begin{array}{c l}
  1-\epsilon & \textrm{if }a = \underset{a_i \in A(s_t)}{argmax}~Q(s_t, a_i)\\
  \epsilon & \textrm{otherwise}
\end{array}
\right.
\end{equation}

Although RL is a plausible learning approach, the agent has to interact with complex state spaces in many situations.
This leads to excessive computational cost in order to find the optimal policy and fully autonomous learning becomes impractical~\cite{scale, moreira2020deep}.

Interactive reinforcement learning (IntRL) is a field of RL research in which a trainer interacts with an RL agent in real-time~\cite{thomaz2005real}.  
IntRL includes an external trainer $\tau$ as an expert to provide advice $\lambda$ for the learning agent in certain situations~\cite{thomaz2005real}.
In this regard, IntRL has been proven as an effective method to speed up the learning process for an artificial agent~\cite{ayala2019reinforcement, millan2019human}.
The advice $\lambda$ provided by the expert trainer may be either evaluative or informative, i.e., it judges the last action performed by the agent or it suggests an action to perform next, respectively~\cite{pilarski2012between}.
Current IntRL methods limit guidance and evaluation to the current state of the agent, regardless of whether the conditions for the information are shared among multiple states~\cite{bignold2020conceptual, cruz2017agent}. 
Therefore, in a particular time step during the learning process, the IntRL agent makes use of the received advice $\lambda_t$ only in that situation, i.e., only in state $s_t$.
After using advice $\lambda_t$ at time step $t$, the agent disregard the advice, not making persistent use of it in case of facing the same situation or a similar one in the future.
This constraint requires the advising user to constantly monitor the current state of the agent and wait until conditions that suit the advice they wish to provide are met again. 
This lack of generalisation increases the number of interactions and the demand on the user~\cite{torrey2013teaching, cruz2017agent}.

In this work, we propose using the advice $\lambda_t$ persistently during the learning process when facing the same state $s$ in a different time step.
Furthermore, we use the same provided advice as a rule for similar situations, i.e., when the agent is in any state $s \in S$ where the rule may be used, then the advice is reused.

\subsection{Rule-based Learning}

In computer science, a rule is a statement consisting of a condition and a conclusion. 
A simple example of a rule is \textit{`IF p THEN q'}, dictating that if the condition of \textit{p} is met, then the conclusion is \textit{q}. 
Additional qualifiers may supplement rules, allowing for a rules condition or conclusion to be constructed to meet specific demands. 
When teaching or conveying information between people, one form of knowledge transfer is rules. 
While the syntax of the rule is not necessarily formal, the relation of condition and conclusion is maintained. 
Moreover, conditions and conclusions are quickly identifiable by humans when natural language is used. 
Recent advances in speech-to-text systems have demonstrated the ability to identify the condition and conclusion in human speech~\cite{lopez2017alexa}. 
The ease with which humans can identify rules for knowledge transfer, and the ability for machines to translate speech to rules, means that rules are an increasingly viable option for knowledge transfer for non-technical users~\cite{churamani2020icub}.

A user may create multiple rules over the duration of their assistance to an agent~\cite{kwok1990multiple}. 
As a result, a single state may have multiple rules, each with conflicting advice for the current state. 
In this regard, binary decision trees offer a method of structuring rules in such a way that only one conclusion is given for each state~\cite{rokach2005decision}.  
Algorithms such as ID3~\cite{quinlan1986induction} and CART~\cite{breiman2017classification} allow the design of the decision tree to be automated, provided that large amounts of labelled data are available. 

The usual methods for building decision trees do not meet the IntRL constraints. 
IntRL does not have access to large amounts of labelled data and aims to be within the skill level of non-expert users, not specialised knowledge engineers~\cite{moreira2020deep}.  
Rule-based IntRL requires a method for generating binary decision trees without the need for expert skills in knowledge engineering, without large amounts of labelled data, and that can be built iteratively without the need for the user to know the full context of the tree.

In relational RL, a combination between RL and inductive logic programming, logical decision trees have been used~\cite{dvzeroski2001relational, li2020towards}. 
An important difference with classical decision trees is that logical decision trees use a relational database or knowledge base to describe a set of facts.
However, a key issue in this representation is the background knowledge needed to use inductive logic programming.
Additionally, another problem that does not benefit the use of algorithms such as Q-learning with relational function abstraction is the nature of Q-values.
Q-values encode both the distance to and the size of the next reward, this becomes especially hard to predict in stochastic and highly complex tasks~\cite{tadepalli2004relational}.


Case-based reasoning (CBR) has been also combined with RL to accelerate learning by making use of heuristic information~\cite{glatt2020decaf}.
These approaches use a heuristic function to choose the next action to be taken.
For instance, the case-based heuristically accelerated RL (CB-HARL) algorithm proposes the reuse of previously learned policies using CBR~\cite{bianchi2009improving}.
In CB-HARL, previous to the action selection, the case similarity is computed based on the current state and the cost of adapting these cases.
The use of heuristics from a base of cases has also led to the development of transfer learning approaches in machine learning~\cite{taylor2009transfer, bianchi2015transferring}.
These methods also deserve attention since they may be converted into interactive methods with straightforward adaptations.


Ripple-down rules (RDR) is a well-known iterative technique for building and maintaining binary decision trees \cite{kang1995multiple, compton1991ripple}. 
RDR is a combination of decision trees and case-based reasoning~\cite{herbert2018intelligent}. 
A case is a collection of potentially relevant material that the system uses to make a classification and is equivalent to the concept of states in RL. 
Each node in an RDR system contains a rule, a classification, and a case. 
The case paired with each node is referred to as the `cornerstone case' and justifies the node's creation~\cite{richards2009two}.

RDR systems require the user to only consider the difference between the current case and the cornerstone case~\cite{richards2009two}. 
Using this methodology, the user does not need to know the context of the entire system, or how new rules will impact its structure. 
The iterative nature of RDR also negates the need for large amounts of labelled data. 
Instead, the tree is built using the gradual flow of cases that any decision tree system is subject to. 
These features make RDR suitable to structure rule-based advice in IntRL scenarios.

\section{A Persistent Rule-based\\Interactive Approach}

The method for retention and reuse of advice proposed here combines the concept of modelling demonstrations, with the evaluative and informative interaction methodology from IntRL 
\cite{bignold2020human}. 
This combination, resulting in retained advice, allows an agent to maximise the utility of each interaction. 
Additionally, a rule-based IntRL approach would further minimise the advisor demand. 
Rule-structured advice allows information to be generalised over multiple states. 
This reduces the interactions required with the human advisor while simultaneously increasing the potential benefit each interaction has on the agent's behaviour. 
The generalisation occurs because the user can specify the conditions in which the information is applicable, allowing the advice to be generalised beyond the current state. 
The agent can then check each state it encounters to see if the conditions are met, at which point the recommendation or evaluation can be utilised.

\subsection{Persistent State-based Interaction}
As introduced, we propose a persistent agent that keeps a record of each interaction and the conditions in which it occurred. 
When the conditions are met in the future, the interaction is replayed. 
This results in improved utilisation of the advice and consequently, improved performance of the agent. 
Additionally, fewer interactions with the trainer are required, as there is no need for advice to be repeatedly provided for each state. 

However, a naive implementation of persistence can introduce flaws into the reward-shaping process. 
These flaws, if unaddressed, may cause the agent to never learn an optimal policy. 
Prior work on reward-shaping~\cite{bicycle} has shown that while reward-shaping can accelerate learning, it can also result in the optimal policy under the shaping reward differing from that which would be optimal without shaping. 
Ng \textit{et al.}~\cite{ng1999policy} demonstrated that this issue can be avoided by using a potential-based approach to constructing the shaping reward signal. 
This guarantees that the rewards obtained along any path from a state back to itself are zero-sum so that the agent will not find a loop in the environment that will provide infinitely accumulating rewards without termination~\cite{devlin2011theoretical}.
For non-persistent IntRL agents, the reward given as part of the evaluation is temporary as the human has to provide the supplemental reward upon revisiting the state. 
Assuming that the human will eventually stop providing advice, the reward signal will become zero-sum~\cite{potentialbasedshaping}. 

For IntRL agents that use policy-shaping, i.e., recommendations on which action to perform next, a straightforward implementation of persistence will work if the advice is correct. 
However, human advice is rarely 100\% accurate~\cite{bignold2020human}. 
Inaccuracy can result from negligence, misunderstanding, latency, maliciousness, and noise introduced when interpreting advice. 
Furthermore, if the agent always performs the recommended action, then it is not given the opportunity to explore and discover the optimal action. 
An agent that retains and reuses inaccurate advice will not learn an optimal policy. 
Therefore, it is important that the agent be able to discard or ignore retained knowledge.

These two issues with persistence, non-potential reward-shaping and incorrect policy-shaping advice, result in persistent agents being unable to learn the optimal policy. 
The issue of inaccurate advice with persistence has two possible solutions, either identify the incorrect advice and discard it or discard all advice after a period regardless of its accuracy. 
To know the accuracy of a piece of advice a full solution to the problem must be known, and if this is achievable then an RL agent is not needed. 
Instead, a policy of discarding or ignoring advice after a period allows a persistent agent to function with potentially inaccurate advice, while still maximising the utility of each interaction.  
This method also solves the issue of non-potential evaluative advice, as the frequency of the supplemental reward is reduced over time until zero. 
Once the supplemental reward is reduced to zero, the cumulative shaping reward function becomes zero-sum once again.  

Therefore, to solve the issue of incorrect advice in persistent IntRL, a method for discarding or ignoring advice after a period of time is needed. 
Probabilistic policy reuse (PPR) is a technique that aims to improve RL agents that use guidance~\cite{fernandez2006probabilistic}. 
PPR relies on using probabilistic bias to determine which exploration policy to use when multiple options are available, the goal of which is to balance random exploration, the use of a guidance policy, and the use of the current policy. 

For the persistent agent scenario, there are three action selection options available: random exploration, the use of retained advice from the trainer, or the best action currently known. 
PPR assigns each of the three options a probability and priority of selection~\cite{fernandez2006probabilistic}. 
Over time, the probability of using guidance or retained information decreases, and trust in the agents own policy increases.
Using PPR, the guidance provided by the trainer is used for more than a single time step, with a decreasing probability over time, until the value of the advice is captured by the agent's own policy. 
Once encapsulated by the agent, self-guided exploration and exploitation of the environment continue.

\subsection{Rule-based Interaction}
Following, we supply details about the rule-based interactive agent implemented. 
As in the previous case, issues of conflicting and incorrect advice need to be mitigated.
Therefore, a method for managing and correcting retained information is required. In this regard, ripple-down rules (RDR) offer a methodology for iteratively building knowledge-based systems without the need for engineering skills. 

While current IntRL agents accept advice pertaining to the current state only, ripple-down rule reinforcement learning (RDR-RL) accepts rule-based advice that can apply to multiple states. 
Each interaction contains a recommendation or evaluation from the user and the conditions for its application. 
For example, the user may provide the following rule to an agent learning to drive a car: \textit{``IF obstacle\textunderscore on left==TRUE THEN action=turn\textunderscore right''}.
In this example, the advice is to turn right, and the condition for its use is that there is an obstacle on the left-hand side of the car. 
While rule-based IntRL assumes that all interactions contain a rule, this rule does not have to be sourced directly from the user. 
The method in which the user interacts with the agent can be by any means, as long as the advice collected results in a set of conditions and the recommendation. 
The user may provide the set of conditions for the applicability of the advice directly, or optionally, the conditions may be discovered using assistive technologies such as case-based reasoning or speech-to-text. 

An RDR-RL agent has three aspects to be considered during its construction, each of which is described in the following sections. 
These aspects are advice gathering, advice modelling, and advice utility.

\subsubsection{Advice Gathering}
The RDR-RL agent has the same foundation as any RL agent. 
The ability to retain and use the advice provided by the user is an addition to the RL agent, built around the existing algorithm. 
Like existing IntRL agents, when no advice has been provided to the agent, it will operate to the exact same as a standard RL agent. 


For instance, at any point during the agent's learning, a user may assist the agent by recommending an action to take. 
When the user begins an interaction, they are provided with the agent's current state, and if available, the current intended action. 
If the user agrees with the intended action the agent presented, or if the user is no longer available, the agent continues learning on its own. 

If the user disagrees with the action the agent is proposing, or if there is no action proposed, then the interaction continues. 
The user is provided with a cornerstone case. 
The cornerstone case is the state in which the user recommended the action that the agent is intending to take. 
The differences between the cornerstone case and the current state are presented to the user. 
If there is no cornerstone case, for example, when it is the first time the user is providing advice to the agent, then only the current state is provided. 
The user recommends an action for the agent to take and creates a rule that distinguishes the two cases, setting the conditions for their recommended action.  
Once the recommended action has been provided, and the rule setting the conditions for its use determined, they are passed to the agent. 
The agent then uses the rule and recommendation to update its model of advice. 

\subsubsection{Advice Modelling}

Advice modelling is the process of storing the information received from the user. 
The agent receives a rule and a recommendation from the user each time an interaction occurs. 
The rule dictates the conditions that must be met for the recommendation to be provided to the agent. 

For instance, using persistence for state-based IntRL may maintain a lookup table for each state and the corresponding recommendation/evaluation that had been provided. 
As we will describe along with the experimental results, this simple method for advice modelling improves performance compared to agents that do not retain advice. 
However, this lookup model does not generalise advice across multiple states and may present difficulties with incorrect advice. 

For rule-based advice, a ripple-down rules decision tree is used to model the advice provided by the user. 
This system allows a model of advice to be iteratively built over time, as the user provides more information to the agent. 
The RDR model is part of the learning agent but is independent of the Q-value policy. 
It is used to assist in action selection. 

When an interaction with the user occurs, the agent is provided with an action recommendation, and a rule governing its use. 
To update the model of advice, the agent provides the current state as a case to the RDR system. 
The system returns a classification node and an insertion node. 
The classification node contains the recommended action based on the advice collected prior to the current state; the recommendation that the user disagrees with given the current state. 
The insertion node is the last node in the branch of the RDR tree that evaluated the current state and is the point at which the new rule will be inserted. 
A new node is created using the rule and recommendation from the user, along with the current state as the cornerstone case. 
If the rule in the insertion node is evaluated TRUE using the information in the current state, then the new node will be inserted as a TRUE child, otherwise, it will be inserted as a FALSE child. 

\subsubsection{Advice Utility}

The last aspect of the agent's construction details when the advice gathered from the user is used by the agent. 
In the previous section, the concept of persistence was discussed. 
There, it was identified as an issue the decreasing agent performance if incorrect advice was provided, or recommended actions were always followed and neglecting exploration. 
To mitigate this issue, PPR was proposed. 

For the RDR-RL agent, the guidance policy is the model of advice. 
The trade-off between exploration and the exploitation of the learned expected-rewards policy continues to be managed by whichever action selection method is preferred by the agent designer. 
For instance, an $\epsilon$-greedy action selection method is used for the experiments in this work. 
In this regard, PPR manages to switch between the action recommended by the advice model and the $\epsilon$-greedy action selection method. 

At each time step, the advising user has a chance to interact with the agent. 
If an interaction occurs, the model is updated. 
When a user first recommends an action, it is expected that the agent will perform it. 
For this reason, the recommended action is always performed on the time step at which it was recommended, regardless of the probabilities currently set by PPR. 

When an agent is selecting an action in a time step where the user has not recommended a previous action, then PPR is used. 
First, the agent's model of advice is checked to see if any advice pertains to the current state. 
If the model recommends an action, then that action is taken with a probability determined by the PPR selection policy. 
If no action is recommended, then the agent's default action selection policy is used, e.g., $\epsilon$-greedy. 

\section{Experimental Environments}

\subsection{Mountain Car}
The mountain car is a control problem in which a car is located in a unidimensional track between two steep hills.
This environment is a well-know benchmark in RL community, therefore, it is a good candidate to initially test our proposed approach.

The car starts at a random position to the bottom of the valley ($-0.6<x<0.4$) with no velocity ($v=0$).
The aim is to reach the top of the right hill.
However, the car engine does not have enough power to claim to the top directly and, therefore, needs to build momentum moving toward the left hill first.

An RL agent controlling the car movements observes two state variables, namely, the position $x$ and the velocity $v$.
The position $x$ varies between -1.2 and 0.6 in the x-axis and the velocity $v$  between -0.07 and 0.07.
The agent can take three different actions: accelerate the car to the left, accelerate the car to the right, and do nothing.

The agent receives a negative reward of $r=-1$ each time step, while no reward is given if a hill is reached ($r=0$).
The learning episode finishes in case the top of the right hill is climbed ($x=0.6$) or after 1,000 iterations in which case the episode is forcibly terminated.

\subsection{Self-driving Car}
\label{c4:self_driving_car}
The self-driving car environment is a control problem in which a simulated car, controlled by the agent, must navigate an environment while avoiding collisions and maximising speed. 
The car has collision sensors positioned around it which can detect if an obstacle is in that position, but not the distance to that position. 
Additionally, the car can observe its current velocity. 
%
All observations made by the agent come from its reference point, this includes the obstacles (e.g., there is an obstacle on my left) and the car's current speed. 
The agent cannot observe its position in the environment. 

Each step, the environment provides the agent reward equal to its current velocity. 
A penalty of -100 is awarded each time that the agent collides with an obstacle. 
Along with the penalty reward, the agent's position resets to a safe position within the map, velocity resets to the lower limit, and the direction of travel is set to face the direction with the longest distance to an obstacle. 

\begin{figure}
\centering
\subfloat[Simulated self-driving car.]{\includegraphics[width=0.5\linewidth]{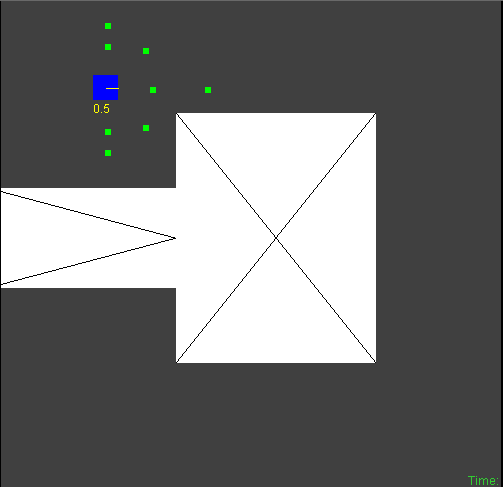} 
\label{c4:fig:sim_car_environment}} 
\subfloat[Optimal path.]{\includegraphics[width=0.5\linewidth]{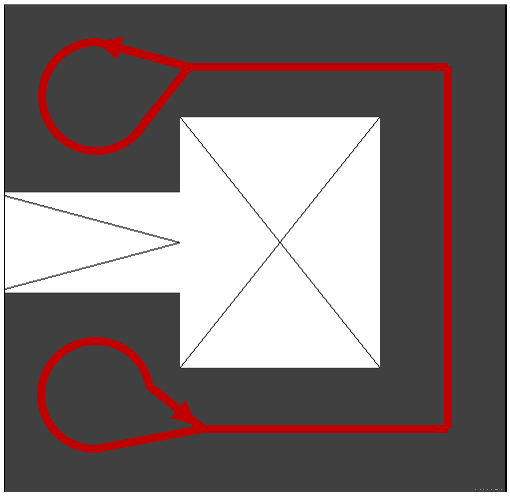} 
\label{c4:fig:sim_car_environment_solution}}
\caption{A graphical representation of the simulated self-driving car. 
The blue square at the top left is the car. 
A yellow line within the car indicates the current direction and the number below (in yellow) is the current velocity. 
The small green squares surrounding the car are collision sensors and will always align with the cars current direction. 
The large white rectangles are obstacles.}
\label{fig:sim_car_environment}
\end{figure}

Figure~\ref{c4:fig:sim_car_environment} shows the map used for the self-driving car experiments. 
This map challenges the agent to learn a behaviour that maximises velocity while avoiding collisions by using a layout that prohibits turning at high speeds at the narrow corridors on the top, right, and bottom of the map. 
The only two sections of the map that allow for high-velocity turning are the large empty sections on the left side. 


The collision sensors return a Boolean response as to whether there is an obstacle at that position, though not the distance to that obstacle. 
Additionally, the agent does not know the position of its sensors in reference to itself. 
The only information the agent has regarding the sensors is whether each is currently colliding with an obstacle. 
As stated, the agent also knows its current velocity. 
The possible velocity of the agent is capped at 1m/s at the lower end, and 5m/s at the higher end. 
A lower cap above zero velocity prevents the agent from moving in reverse or standing still. 
This lower limit reduces the state space and prevents an unintended solution, e.g., standing still is an excellent method for avoiding collisions.  
The upper limit of 5m/s is set so that velocity is not limitless and further reduces the state space, while still being high enough that it exceeds the limit for a safe turn anywhere in the environment. 
An action that attempts to exceed the velocity thresholds set by the environment will return the respective limit. 
%
%
There are five possible actions for the agent to take within the self-driving car environment. 
These actions are:

\begin{enumerate}

\item[i.] \textit{Accelerate}: the car increases its velocity by 0.5 meters per second. 

\item[ii.] \textit{Decelerate}: 
the car's velocity will decrease by 0.5 meters per second. 

\item[iii.] \textit{Turn left}: the car alters its direction of travel by 5 degrees to the left. 

\item[iv.] \textit{Turn right}: the car alters its direction of travel by 5 degrees to the right.

\item[v.] \textit{Do nothing}: the car's velocity or direction of travel is not altered. 
When performing this action the only change is the car's position, based on current velocity, position, and direction of travel. 
\end{enumerate}

The self-driving car environment has nine state features, one for each of the collision sensors on the car, and the current velocity of the car. 
The collision sensor state features are Boolean, representing whether they detect an obstacle at their position. 
The velocity of the agent has nine possible values, the upper and lower limits, plus every increment of 0.5 value in between. 
With the inclusion of the five possible actions, this environment has $11520$ state-action pairs.

The reward function defined by the environment promotes the agent to learn a behaviour that avoids obstacles while attempting to achieve the highest velocity the environment allows. 
The most natural solution to learn that achieves these conditions is to drive in a circle, assuming that the path of the circle does not intersect with an obstacle. 
The map chosen for use in these experiments allows an unobstructed circle path to be found, but only at low velocities. 
If the agent is to meet both conditions that achieve the highest reward, a more complex behaviour must be learned (see Figure~\ref{c4:fig:sim_car_environment_solution}). 


\section{Experimental Methodology}

To compare agent performance and interaction, metrics for agent steps, agent reward, and interactions are recorded. 
A number of different agents and simulated users have been designed and applied to the mountain car and self-driving car environments. 
Simulated users have been chosen over actual human trials, as they allow rapid and controlled experiments~\cite{bignold2021evaluation}. 
When employing simulated users, interaction characteristics such as knowledge level, accuracy, and availability can be set to specific and measurable levels. 
Following, we describe all the agents used during the experiments.

\subsection{Non-Persistent and Persistent\\State-based Agents}

Next, we demonstrate the use of persistent advice with probabilistic policy reuse (PPR), and the impact its use has on agent performance and user reliance. 
The experiments have been designed to test several levels of human advice accuracy and availability, with and without retention of received advice. 

The mountain car environment is used in these experiments since it is a common benchmark problem in RL with sufficient complexity to effectively test agents and simple enough for human observers to intuitively calculate the correct policy. 
Additionally, the mountain car environment has been previously used in a human trial evaluating different advice delivery styles~\cite{bignold2020human} and with simulated user~\cite{bignold2021evaluation}.
We use the results reported in the human trial to set a realistic level of interaction for evaluative and informative advice agents. 
Five agents have been designed for the following experiments.
The expected-reward values have been initialized to zero, an optimistic value for the environment.  
All the agents are given a learning rate $\alpha=0.25$, a discounting factor $\gamma=0.9$, and use an $\epsilon$-greedy action selection strategy with $\epsilon=0.1$.
For the agent to represent the continuous two-dimensional state space of the environment, it has been discretized into 20 bins for each state feature, creating a total of 400 states, each with three actions. 
The learning agents are listed below:

\begin{enumerate}

\item[i.] \textit{Unassisted Q-Learning Agent}: A Q-learning agent used for capturing a baseline for performance on the mountain car environment. 
This agent is unassisted, receiving no guidance or evaluation from the trainer and used as a benchmark. 

\item[ii.] \textit{Non-Persistent Evaluative Advice Agent}: This agent is assisted by a user. 
The user may provide an additional reward at each time step to evaluate the agent's last choice of action. 
For this non-persistent agent, the supplemental reward is used in the current learning step and then discarded. 

\item[iii.] \textit{Persistent Evaluative Advice Agent}: This agent is assisted by a user. 
The user may provide an additional reward at each time step to evaluate the agent's last choice of action. 
For this persistent agent, the evaluation provided is retained, and upon performing the same state-action pair in the future, the evaluation may be automatically provided to the agent, with a probability defined by the PPR action selection policy. 

\item[iv.] \textit{Non-Persistent Informative Advice Agent}: This agent is assisted by a user. 
The user may recommend an action for the agent to perform for the current time step. 
When the agent is recommended an action, that action is taken on that time step, and then the advice is discarded. 
This non-persistent agent, when visiting the same state again in the future, will not recall the recommended action and will perform $\epsilon$-greedy action selection. 

\item[v.] \textit{Persistent Informative Advice Agent}: This agent is assisted by a user. 
The user may recommend an action each time step for the agent to perform. 
If recommended, the learning agent will take the advice on that time step and retain the recommendation for use when it visits the same state in the future. 
When the agent visits a state in which it was previously recommended, it will take that action with the probability defined by the PPR action selection policy. 

\end{enumerate}

\begin{figure}
\centering
\includegraphics[width=1\linewidth]{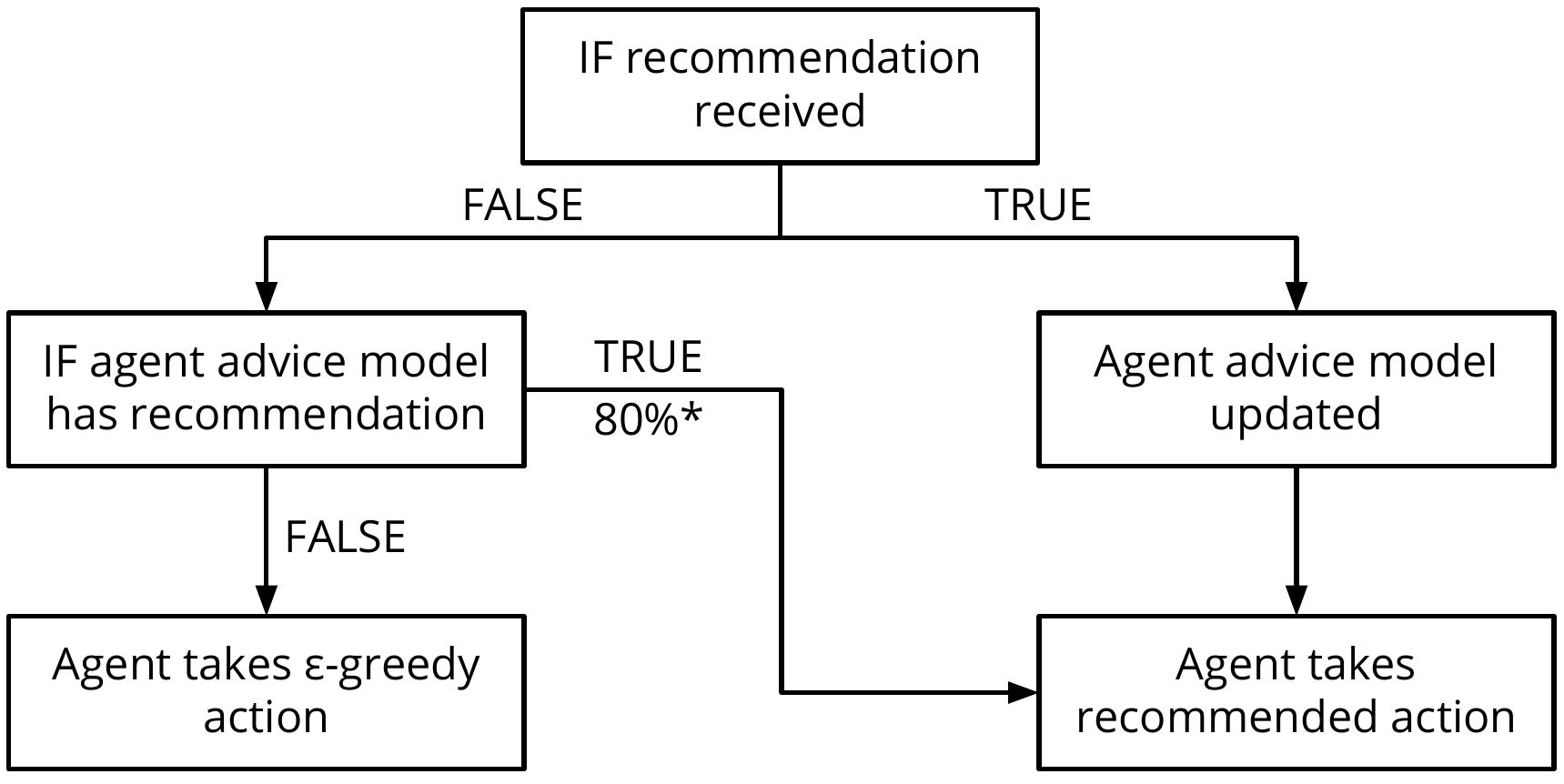}
\caption{Probabilistic policy reuse (PPR) for an IntRL agent using informative advice. If the user recommends an action on the current time step then the agent's advice model updates and the action is performed. If the user does not provide advice on the current time step, then the agent will follow previously obtained advice 80\% of the time (*decays over time) and its default exploration policy the remaining time.}
\label{fig:c7_ppr}
\end{figure}

The agents adopting a persistent model are employing PPR for action selection. 
As depicted in Figure~\ref{fig:c7_ppr}, the PPR action selection begins with an 80\% chance of reusing advice provided to the agent in the past. 
The probability of reusing advice decreases by 5\% each episode. 
For the remaining 20\% of the time, or if no advice has been provided for the current state, an $\epsilon$-greedy action selection policy is used. 

For each agent, one hundred experiments are run. 
At the beginning of each experiment, the environment, the agent, and the agent's model of provided advice are reset. 
Each experiment runs with a maximum of one thousand steps before it terminates. 
The number of steps performed, interactions performed, and reward received are recorded. 
An interaction is recorded if the user provides advice to the agent, not when the agent uses advice it has stored from a previous interaction. 

Six different simulated users have been created as trainers. 
Three providing evaluative advice and three informative advice. 
Evaluative advice-giving users provide either a positive or negative supplemental reward corresponding to the agent's choice in action on the last time step. 
Informative advice-giving users provide a recommended action for the agent to perform on the current time step. 
Simulated users that are advising a persistent agent will not provide advice for a state, or state-action, that it has previously advised on, as it is assumed that if the agent is retaining information it should not need repeated advice for the same conditions. 
This does not apply to non-persistent agents.

Additionally, each simulated user will have either optimistic, realistic, or pessimistic values for advice accuracy and availability. 
Accuracy is a measure of how correct the advice is provided by the user. 
Accuracy of interaction is altered by, with a specified probability, replacing the recommended action with an action that is not optimal for the current state. 
Availability is a measure of how frequently the user provides advice. 
The availability of the simulated user is altered by specifying a probability that the user will interact with the agent on any given time step.

Optimistic simulated users have 100\% accurate advice and will provide advice on every time step that the agent does not have retained knowledge of. 
Realistic simulated users have accuracy and availability modelled from previously obtained results in a human trial~\cite{bignold2020human}. 
The recorded accuracy and availability of human advice-givers differs depending on the type of advice being provided, i.e., evaluative or informative. 
Previous work has compared evaluative and informative advice/agents~\cite{bignold2020human}, and as such is not in the scope of this study. 
Lastly, pessimistic simulated users are given accuracy and availability values half that of the realistic users. 
Table~\ref{c7:tab:sim_user_settings} shows the accuracy and availability values for each of the six simulated users (3 evaluative users, 3 informative users). 
The previously observed accuracy and availability for human advisors in the mountain car environment are shown as for realistic agents.

\begin{table}
\caption{Simulated users modelled for the experimental setup. 
Accuracy and availability are set using previous results obtained in a human trial as reference~\cite{bignold2020human}.
}

\begin{center}
\begin{tabular}{|l|l|l|}
 \hline
 \textbf{Name}& \textbf{Acc.} & \textbf{Avail.} \\
 \hline
 Evaluative Optimistic		   	& 100\%    & 100\%       \\
 Evaluative Realistic			& 48.470\% & 26.860\%    \\
 Evaluative Pessimistic 		& 24.235\% & 13.43\%     \\
 \hline
 Informative Optimistic   		& 100\%    & 100\%       \\
 Informative Realistic			& 94.870\% & 47.316\%    \\
 Informative Pessimistic		& 47.435\% & 23.658\%    \\
 \hline
\end{tabular}
\end{center}
\label{c7:tab:sim_user_settings}
\end{table}

Table~\ref{c7:tab:agent_user_combinations} lists all the agent/simulated user combinations tested. 
There are a total of thirteen agents, six persistent agents, six non-persistent agents, and an unassisted Q-Learning agent used for benchmarking. 
Included next to each agent/user combination is a short name. 
This short name is used later in the Results section, as the full name is too long to include in each figure legend.

\begin{table}[t!]
\caption{Agent/User combinations for persistent agent testing, including short names for reference.}
\begin{center}
\begin{tabular}{ |p{1.2cm}|p{2.5cm}|p{2.9cm}|  }
 \hline
 \textbf{Short Name} & \textbf{Agent} & \textbf{Simulated User} \\
 \hline
 UQL   & Unassisted Q-Learning    	& NONE\\
 \hline
 NPE-O & Non-Persistent Evaluative		& EVALUATIVE OPTIMISTIC\\
 NPE-R & Non-Persistent Evaluative		& EVALUATIVE REALISTIC\\
 NPE-P & Non-Persistent Evaluative		& EVALUATIVE PESSIMISTIC\\
 \hline
 NPI-O & Non-Persistent Informative 	& INFORMATIVE OPTIMISTIC\\
 NPI-R & Non-Persistent Informative 	& INFORMATIVE REALISTIC\\
 NPI-P & Non-Persistent Informative 	& INFORMATIVE PESSIMISTIC\\
 \hline
 PE-O  & Persistent Evaluative 			& EVALUATIVE OPTIMISTIC\\
 PE-R  & Persistent Evaluative 			& EVALUATIVE REALISTIC\\
 PE-P  & Persistent Evaluative 			& EVALUATIVE PESSIMISTIC\\
 \hline
 PI-O  & Persistent Informative 		& INFORMATIVE OPTIMISTIC\\
 PI-R  & Persistent Informative 		& INFORMATIVE REALISTIC\\
 PI-P  & Persistent Informative 		& INFORMATIVE PESSIMISTIC\\
 \hline
\end{tabular}
\end{center}
\label{c7:tab:agent_user_combinations}
\vskip -0.1in
\end{table}

\subsection{Rule-based Agents}
In this case, three learning agents have been designed, which include unassisted Q-Learning, persistent state-based informative, and rule-based assisted using ripple-down rules. 
No evaluative assisted agents are tested in these experiments, as they cannot be suitably compared to the rule-assisted agent which is using informative advice. 
The three learning agents used are described below: 

\begin{enumerate}
\item[i.] \textit{Unassisted Q-Learning Agent}: A Q-learning agent used for capturing a baseline for performance on each environment. 
This agent is unassisted, receiving no guidance or evaluation from the trainer and used as a benchmark. 
The agent will represent each environment as described in the previous section. 
The expected-reward values have been initialized to zero. 
This agent uses $\epsilon$-greedy action selection.

\item[ii.] \textit{State-based Persistent Agent}: This agent is assisted by a user. 
The user may recommend an action each time step for the agent to perform. 
If an action is recommended by the user, the agent will take it on that time step and retain the recommendation for use when it visits the same state in the future. 
When the agent visits a state in which it was previously recommended, it will take that action with the probability defined by the PPR action selection strategy. 
The persistent informative agent uses the same parameter settings as the unassisted Q-Learning agent for each environment.

\item[iii.] \textit{Rule-Assisted Persistent Agent}: This agent is assisted by a user. 
The user may provide a rule and recommended action at each time step. 
The rule-assisted learning agent uses ripple-down rules to model the advice received by the trainer. 
If the user provides advice, and the rule provided equates to true for the current state, then the agent will take the recommended action during that time step. 
If the provided rule equates to false, then the agent will use its default action selection strategy. 
When a rule is provided the agent will retain the rule for use in future states. 
Each time the agent visits a state, it will query its retained model of rules. 
If a rule is found that equates to true for the current state, then that action is taken with a probability defined by the agent's PPR action selection strategy. 
All rule-assisted agents used in this experiment begin with an 80\% chance of taking the action recommended by its advice model. 
This 80\% chance is decayed each episode, until the point at which the agent is relying solely on its secondary action selection strategy. 
The agent's secondary action selection strategy is the same strategy used by the unassisted Q-Learning agent, i.e., $\epsilon$-greedy. 
The rule-assisted agent uses the same parameter settings as the unassisted Q-Learning agent for each environment.

\end{enumerate}

The mountain car environment is a good candidate for the rule-based advice method as the optimal solution can be captured in very few rules, while still remaining understandable by humans. 
The rule-based and state-based agents are tested against the mountain car environment, employing simulated users with varying levels of knowledge of the environment. 
The aim is to compare the performance of the agents, and the number of interactions performed to achieve that performance. 
The learning parameters used are the same as the previous experiments.

Additionally, in these experiments, the self-driving car environment is also used.
The state and action spaces for this environment is larger than the mountain car environment, but still remain understandable by human observers. 
The self-driving car agents are given a learning rate $\alpha=0.1$, a discounting factor $\gamma=0.999$, and used an $\epsilon$-greedy action selection strategy with $\epsilon=0.01$.

The requirements of the reward function, to avoid collisions and to maximise velocity, make the creation of optimal rules much more difficult. 
For the self-driving car environment, it is easy to provide rules that will help achieve greater performance in parts of the environment, maximising speed or when to turn for example. 
However, it is much more difficult to provide rules that meet both requirements optimally, for example, when to turn the car and by how much to maintain the highest possible velocity while not crashing. 
The characteristic of being able to easily creating performance improving yet non-optimal rules is what makes the self-driving car environment an interesting benchmark for the rule-based advice method. 
The difference between this environment and the mountain car is that this environment will test a larger state and feature space, and consist of advice that, while beneficial, is not optimal.

\subsection{Simulated Users}
To allow quick, bias-reduced, repeatable testing of the agents, simulated users are used as trainers in place of humans. 
Simulated users offer a method for performing indicative evaluations of RL agents that require human input, with controlled parameters~\cite{bignold2021evaluation}. 
There are two types of simulated users required for the following experiments, one must provide state-based advice, and the other must provide rule-based advice. 
Both types of simulated users will provide the same information and the same amount of it. 

\begin{table}[t!]
\caption{State-based simulated user knowledge bases for the mountain car and the self-driving car environments.}
\begin{center}
\begin{tabular}{ |p{2.5cm}|p{4.7cm}|  }
 \hline
 \textbf{Environment (User Name)} & \textbf{Limits}\\
 \hline
 Mountain Car / MC-FULL & User will provide advice for all states.\\
 \hline
 Mountain Car / MC-HALF & User will only provide advice for state in which the agent is on the left slope of the valley. (IF position $<-0.53$)\\
 \hline
 Mountain Car / MC-QUARTER & User will only provide advice for state in which the agent is on the bottom half of the left slope of the valley. (IF position $<-0.53$ AND position $>-0.865$)\\
 \hline
 Mountain Car / MC-MIDDLE & User will only provide advice for the few states at the bottom of the valley. (IF position $< -0.43$ AND position $>-0.63$)\\
 \hline
 Self-driving Car / SC-AVOID & User will only provide advice for states where the agent has an obstacle on the left side OR the right side, but not both. (IF right = true OR right-front-close = true) OR (IF left = true OR left-front-close = true)\\
 \hline
\end{tabular}
\label{c8:tab:sim_user_settings_state}
\end{center}
\end{table}

The first type, an informative state-based advice user, is the same user employed for the previous experiments. 
This user may provide a recommended action on each time step. 
The agent that the user is assisting will retain any recommendations provided by the user, and will not give the user an opportunity to provide advice for a state for which advice has already been received, capping the number of interactions at the number of states. 
As in the previous experiments, each informative state-based user had an accuracy and availability score. 
Accuracy is the probability that the advice the user is providing is optimal for the current state. 
Availability is the probability that the user would provide advice for any given opportunity. 
Additionally, the states that the user can provide advice will be limited to parts of the environment, simulating a limited or incomplete knowledge level of the environment. 
Table~\ref{c8:tab:sim_user_settings_state} shows the knowledge limitations of the various state-based users built for the rule-based experiments in order to do a fair comparison.

The advice that the state-based simulated users provide for the mountain car environment is optimal (previous to accuracy, availability, and knowledge level is applied). 
However, the same may not be true for the self-driving car environment. 
The reward function for this environment reinforces behaviour that avoids collisions and maximises speed. 
The advice that the simulated user provides for the self-driving car environment only attempts to avoid collisions. 
While this advice should be optimal, there may be situations where the agent will want to stay close to an obstacle to maximising speed. 
In these situations, the advice provided would be considered incorrect, and the agent will need to learn to ignore it to learn the optimal behaviour. 

The second type of simulated user is a rule-based advice user. 
Simulated users are a common methodology for the creation and evaluation of ripple-down rule systems in research~\cite{kang1995multiple, kang1998simulated, compton1995use}. 
These simulated users will return a rule and a recommended action for each interaction with the user. 
The simulated users employed for the these experiments have been built with their own ripple-down rules model and populated with a set of rules that they will, over time, provide to the agent. 
As in reality, users do not have their own rule model, rather they would generate rules themselves, therefore, we use the rule model for simulated users as a means to replicate the interaction process of a real user.

The learning agent begins each experiment with an empty model of advice, and the simulated user begins with a full model. 
Over time, the learning agent will provide the trainer user with an opportunity to provide advice. 
When an opportunity occurs, the learning agent provides the current state observation, the current action it will take, and details about how it chose that action (either from the retained user model or from an exploration strategy). 
This information is the same information that would be made available to an actual human advisor. 
Now that the simulated user has this information, it may choose to respond and what advice it will provide with. 
The simulated user will respond if it has a rule that applies to the current state and it disagrees with the agent's choice of action. 
The simulated user will continue to provide advice for as long as it is given opportunities, that it has new rules to provide, and that the new rules disagree with the agent's current behaviour. 
Algorithm~\ref{alg:IRL} shows the full process flow to choose an action using rule-based advice to assist a learning agent.
In the algorithm, $c_{t+1}$ represents the cornerstone case for state $s_{t+1}$ and action $a_{t+1}$, whereas $l_{t+1}$ represents the advice given by the user at state $s_{t+1}$.

\begin{algorithm}
\caption{Interactive reinforcement learning with a rule-based advice model for assisting an RL agent.}\label{alg:IRL}
\begin{algorithmic}[1]
\State Initialize environment selecting $s_t$
\For{(each episode)}
  \State Choose action $a_t$ from $s_t$ using $\pi$
  \Repeat 
    \State Perform action $a_t$
    \State Observe next state $s_{t+1}$
    \State Choose next action $a_{t+1}$ from $s_{t+1}$
    \State Pass $s_{t+1}, a_{t+1}, c_{t+1}$ to user
    \If {(adding advice)} 
      \State Observe advice $l_{t+1}$ from $s_{t+1}$
      \If {($l_{t+1} \neq a_{t+1}$)}
        \State Create new rule using $c_{t+1}$
        \State Update advice model
        \State Change action $a_{t+1}$
      \EndIf
    \Else
      \State User ignores agent
    \EndIf
    \If {$rand(0,1) < \epsilon$} 
      \State Choose a random action $a_{t+1}$ from $A$
    \EndIf
    \State Update Q-values
    \State $s_t\leftarrow s_{t+1}; a_t\leftarrow a_{t+1}$
  \Until {$s$ is terminal}
\EndFor
\end{algorithmic}
\end{algorithm}

Multiple rule-based simulated users have been created to provide a range of different knowledge levels for the various environments (equivalent to the knowledge level of state-based simulated users shown in Table~\ref{c8:tab:sim_user_settings_state}). 
Table~\ref{c8:tab:sim_user_settings_rule} describes and provides the knowledge bases in use for each of the environments. 
A short description of each knowledge base is provided.

\begin{table}[t!]
\caption{Rule-based simulated user knowledge bases for the mountain car and self-driving car environments. The model is shown using ripple-down rules with a text representation.}
\begin{center}
\begin{tabular}{ |p{2.24cm}|p{5.4cm}|  }
 \hline
 \textbf{Environment / User Name} & \textbf{Limits}\\
 \hline
 \shortstack[l]{Mountain Car / \\ MC-FULL} & \shortstack[l]{
 IF 1==1 : EXPLORE\\
 \quad IF velocity $>$ 0: GO RIGHT\\
 \quad \quad NO TRUE NODE\\
 \quad \quad IF velocity $<$=0 GO LEFT} \\
 \hline
  \shortstack[l]{Mountain Car / \\ MC-HALF} & \shortstack[l]{
 IF 1==1 : EXPLORE\\
 \quad IF position $<$ -0.53: GO RIGHT\\
 \quad \quad IF velocity $>=$ 0: GO RIGHT\\
 \quad \quad IF velocity $<$ 0: GO LEFT\\
 \quad NO FALSE NODE
 }\\
 \hline
 \shortstack[l]{Mountain Car / \\ MC-QUARTER} & \shortstack[l]{
 IF 1==1 : EXPLORE\\
 \quad IF position $<$ -0.53 AND \\
 \quad \quad position $>$ -0.86: GO RIGHT\\
 \quad \quad IF velocity $>=$ 0: GO RIGHT\\
 \quad \quad IF velocity $<$ 0: GO LEFT\\
 \quad NO FALSE NODE
 }\\
 \hline
 \shortstack[l]{Mountain Car / \\ MC-MIDDLE} & \shortstack[l]{
 IF 1==1 : EXPLORE\\
 \quad IF position $<$ -0.43 AND \\
 \quad \quad position $>$ -0.63: GO RIGHT\\
 \quad \quad IF velocity $>=$ 0: GO RIGHT\\
 \quad \quad IF velocity $<$ 0: GO LEFT\\
 \quad NO FALSE NODE
 }\\
 \hline
 \shortstack[l]{Self-driving Car \\ / SC-AVOID} & \shortstack[l]{
 IF 1==1 : EXPLORE\\
 \quad IF right OR right-front-close:\\
 \quad \quad TURN LEFT\\
 \quad \quad NO TRUE NODE\\
 \quad \quad IF left OR left-front-close:\\
 \quad \quad TURN RIGHT\\
 \quad NO FALSE NODE}\\
 \hline
\end{tabular}
\end{center}

\label{c8:tab:sim_user_settings_rule}
\vskip -0.2in
\end{table}

\section{Results}


\subsection{Probabilistic Policy Reuse}

\begin{figure}
\centering
\includegraphics[width=1\linewidth]{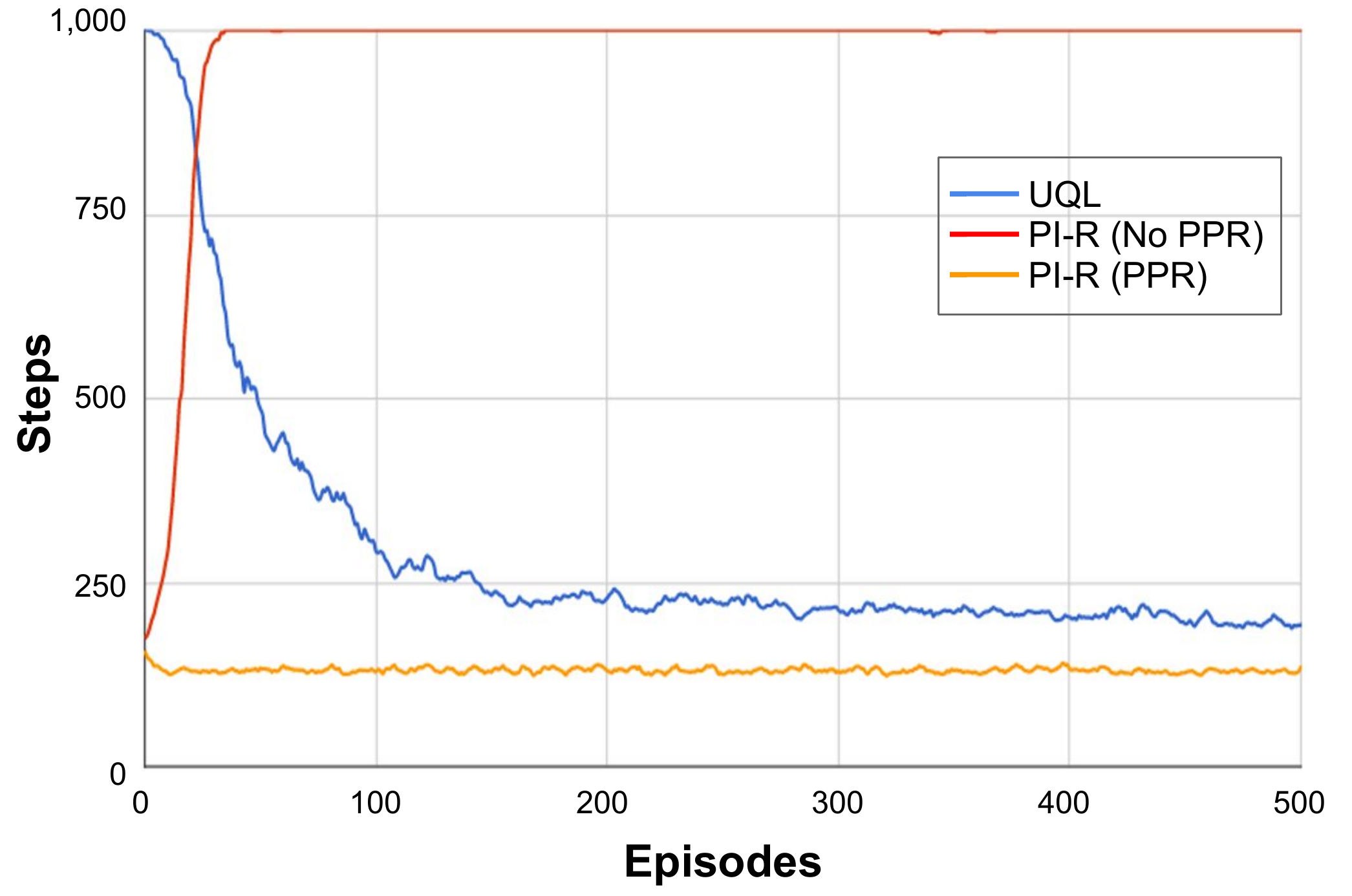}
\caption{Probabilistic policy reuse and direct-use action selection for IntRL using retained informative advice. 
Both assisted agents are using simulated users using realistic values for accuracy and availability, and are both retaining advice provided to them.}
\label{c7:fig:ppr_compare}
\end{figure}

The first experiment performed tests the use of probabilistic policy reuse (PPR) as an action selection method, compared to always using advice when available within the mountain car environment.
As aforementioned, the use of persistence in RL introduces a critical flaw. 
Specifically, if provided advice is retained and reused, and that advice is incorrect, then the agent will not be able to learn a solution to the current problem. 
Figure~\ref{c7:fig:ppr_compare} shows the performance of 3 RL agents: an unassisted Q-learning (UQL) agent for benchmarking and 2 persistent IntRL agents using informative advice. 
These two interactive agents are identical except that one is using PPR for action selection, called persistent informative reuse (PPR) agent or PI-R (PPR), while the other will always take a recommended action if one exists for the current state, called persistent informative reuse (No-PPR) agent or PI-R (No-PPR). 
Both interactive agents are assisted by a simulated user created with realistic values of accuracy and availability.

Figure~\ref{c7:fig:ppr_compare} shows that both assisted agents immediately outperforms an unassisted agent (UQL in blue). 
Both agents are retaining the recommended actions from the user, and cannot differentiate between correct and incorrect advice. 
The No-PPR agent (in red) will always take the recommended action for the current state, if available. 
This works well for the first few episodes, as small amounts of correct advice can have a large positive impact on agent performance and small amounts of incorrect can be ignored because of the momentum the agent builds in the mountain car environment. 
However, as the amount of incorrect recommended actions retained increases, the effect on the agent's performance increases. 
Eventually, the impact of taking the wrong action will have such an effect that the agent cannot build the required momentum to solve the task. 
Without the required momentum, the agent will get stuck in local minima. 

The agent using probabilistic policy reuse (PPR) continues to outperform both the unassisted agent (UQL), and the other assisted agent (PI-R No-PPR). 
The PPR agent will initially take the users advice in high regard, taking recommended actions 80\% of the time. 
Over time, the agent pays less attention to the retained advice of the user, and more to its own learned policy. 
This allows the agent to disregard incorrect advice, as its own value estimations will show the correct action to take, while correct advice will accelerate the discovery of the true value estimation of the correct action in advised states. 

If human-sourced advice is 100\% accurate for the problem being tested, the use of PPR may lower the potential performance of the agent. 
This is due to the PPR action selection policy disregarding accurate information and instead taking exploratory or local minima actions. 
However, previous work~\cite{bignold2020human} has shown that human-sourced information is not likely to be 100\% correct, and as such, the use of PPR mitigates the risk of inaccurate information.

\subsection{State-based Persistent Advice} 

\begin{figure*}[!hbt]
\centering
\subfloat[Non-persistent evaluative]{\includegraphics[width=0.5\linewidth]{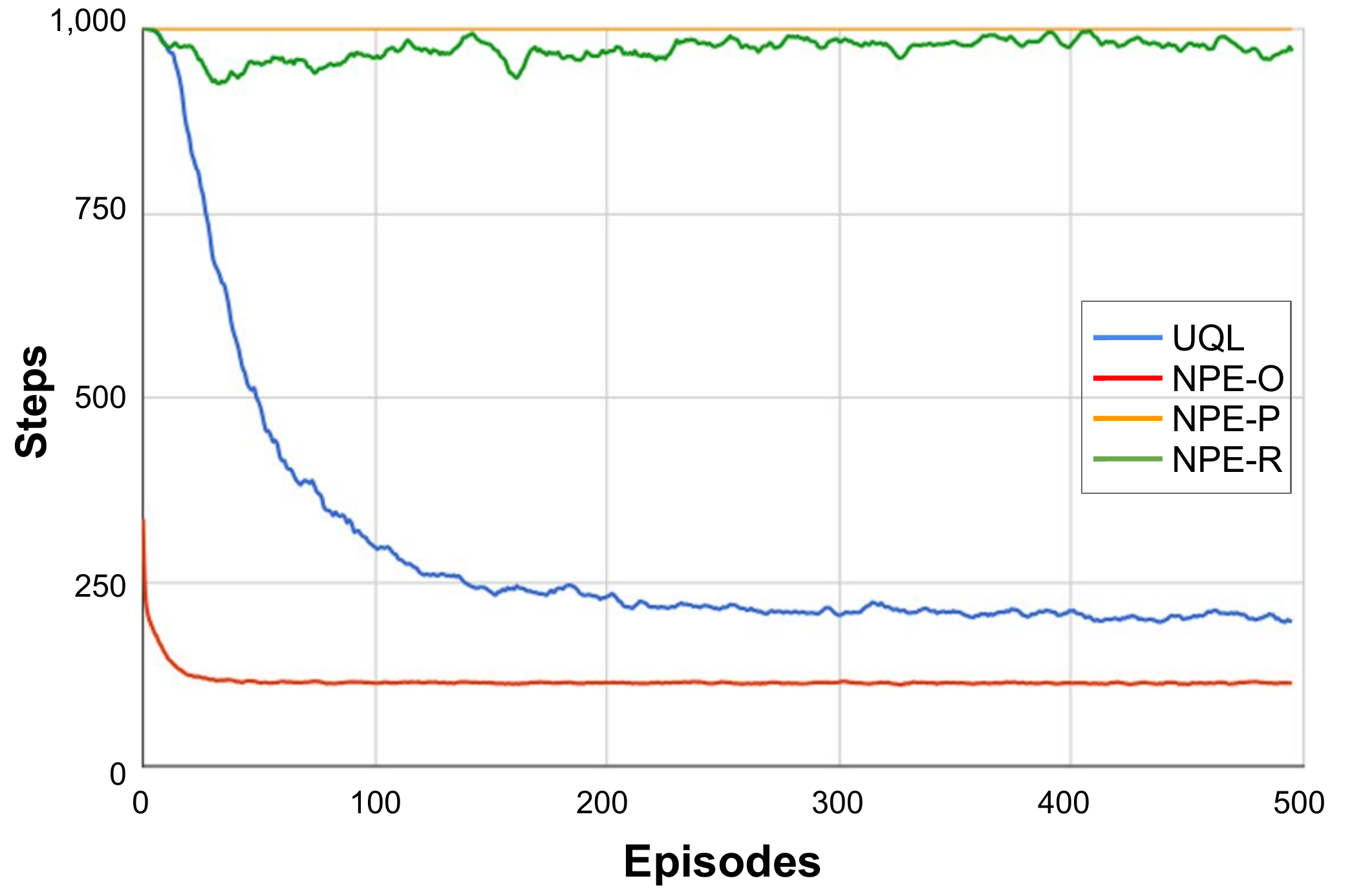}
\label{c7:fig:evaluative_persistent_a}} 
\subfloat[Persistent evaluative]{\includegraphics[width=0.5\linewidth]{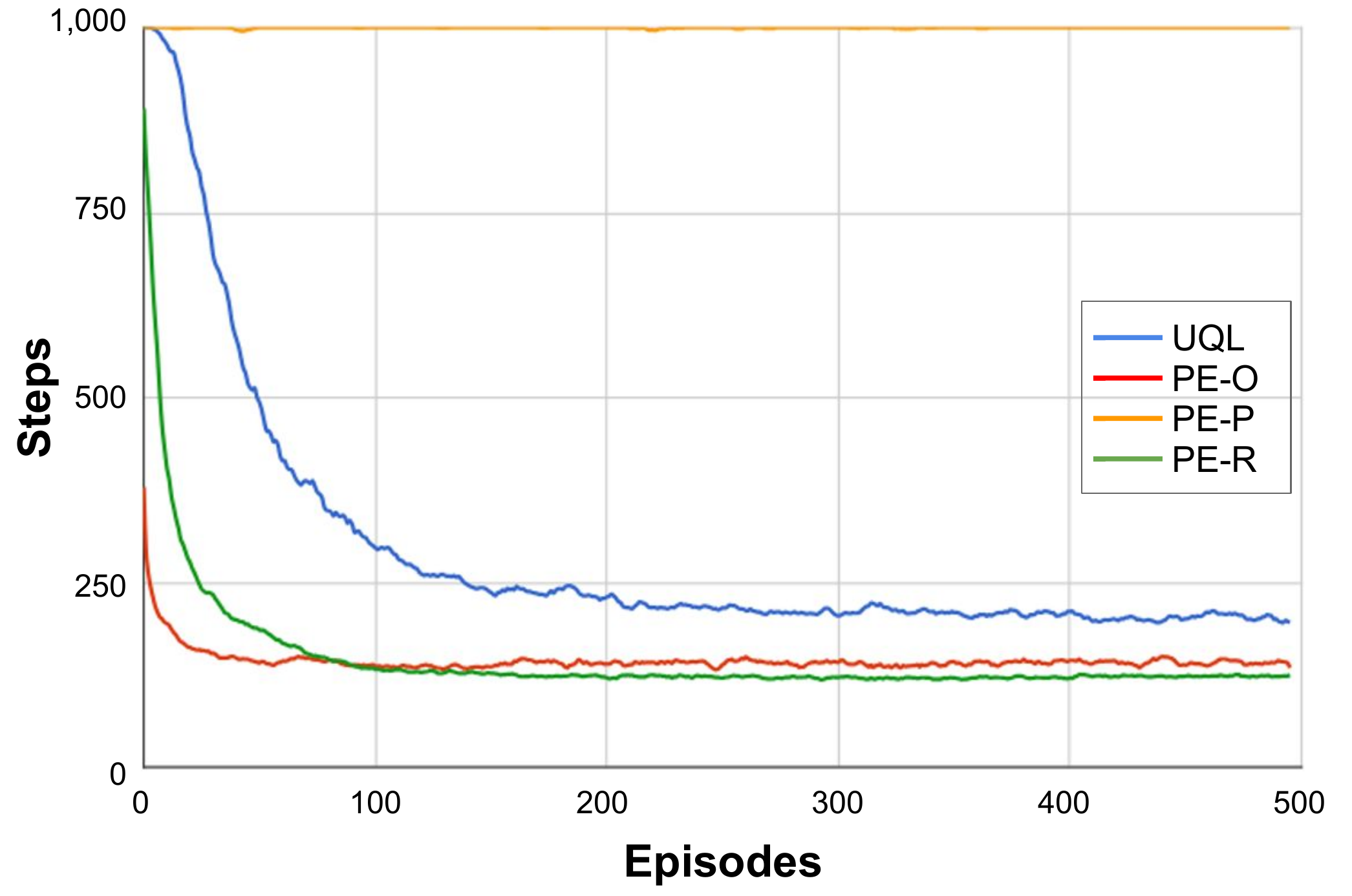}
\label{c7:fig:evaluative_persistent_b}}
\\
\subfloat[Non-persistent informative]{\includegraphics[width=0.5\linewidth]{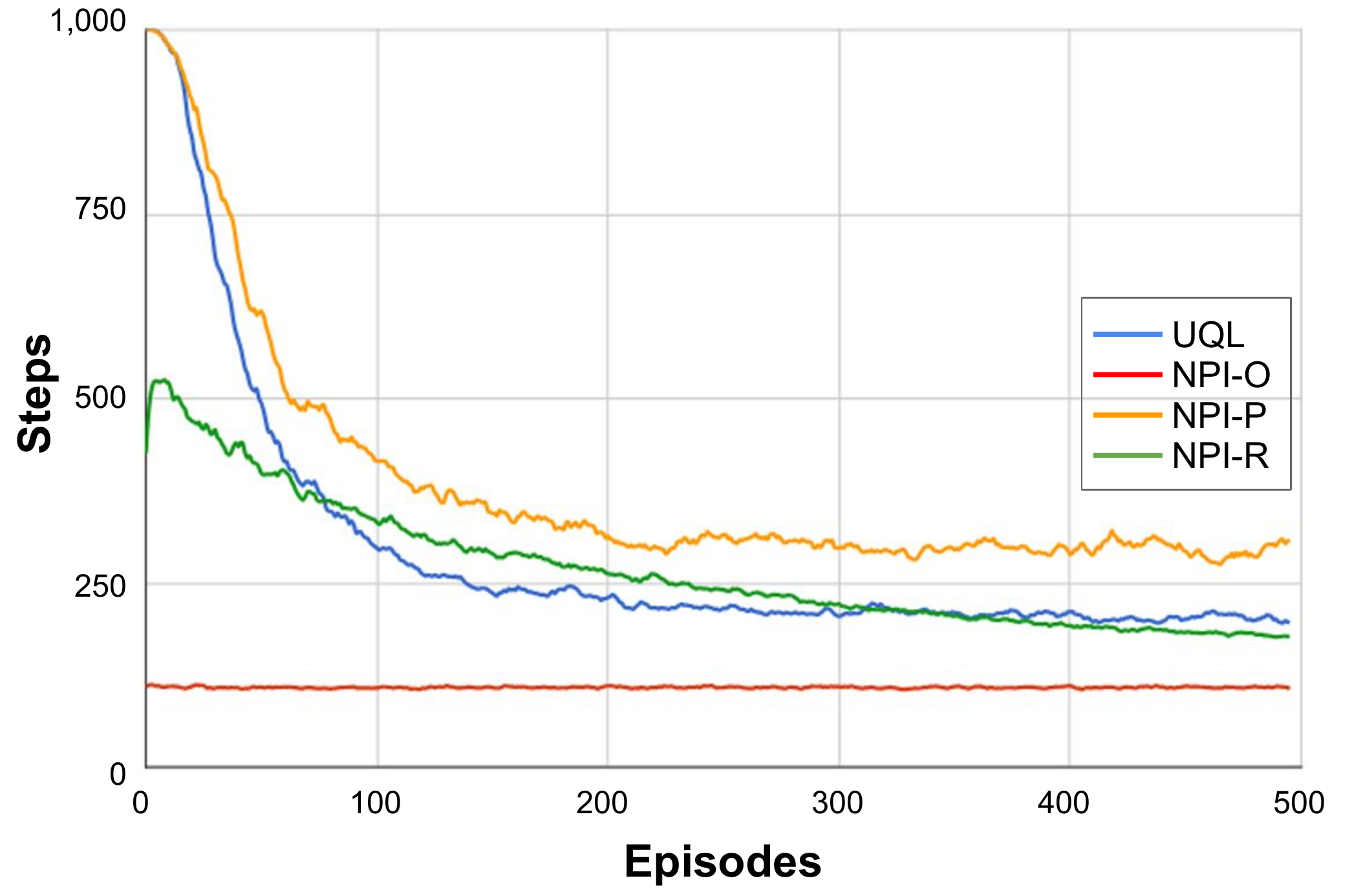}
\label{c7:fig:informative_persistent_c}}
\subfloat[Persistent informative]{\includegraphics[width=0.5\linewidth]{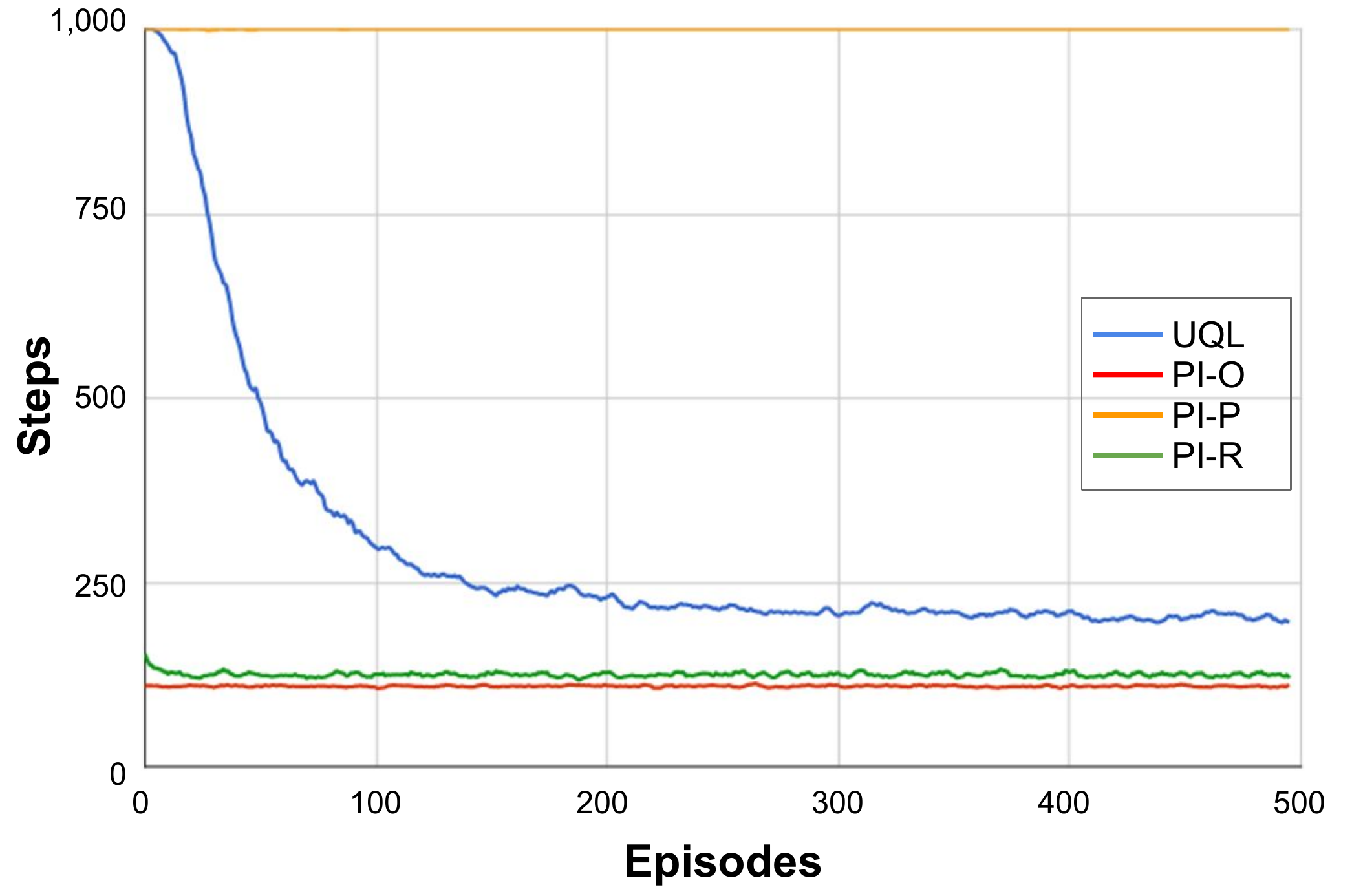}
\label{c7:fig:informative_persistent_d}}
\\
\caption{Steps per episode for 4 different agents using advice. 
The agents are assisted by three different simulated users, initialised with either \textbf{O}ptimistic, \textbf{R}ealistic, or \textbf{P}essimistic values for accuracy and availability. 
The figure shows that the persistent agents learn in fewer steps in comparison to the non-persistent agents when assisted by sufficiently accurate users.
}
\label{fig:state-based}
\end{figure*}

The second experiment tests non-persistent advice and state-based persistent advice, i.e., the provided advice is used and retain with PPR only in the given state.
An unassisted Q-learning (UQL) agent is used for benchmarking. 
Simulated users are used for providing advice in the mountain car environment with three different initialisation, namely, optimistically, realistic, and pessimistically (denoted with the suffix -O, -R, and -P respectively).
Figures~\ref{c7:fig:evaluative_persistent_a} and~\ref{c7:fig:informative_persistent_c} show the performance over time for both non-persistent evaluative and informative agents, at varying levels of user accuracy and availability. 
These figures do not compare the two advice delivery styles against each other (i.e., evaluative and informative), but they are compared against their persistent counterparts. 
As evaluative advice is evaluating actions that have already been taken, there is a short delay between the action being taken and the application of the advice to the agent. 
This delay causes latency in the effect of the advice on the agent's learned policy. 
Figure~\ref{c7:fig:evaluative_persistent_a} shows this delay for the evaluative agent, where most of the advice is given in the first few episodes, but it takes around twenty episodes before the agent has fully utilised the advice and converges to an optimal path. 
The agent using informative advice on the other hand does not suffer from this delay (shown in Figure~\ref{c7:fig:informative_persistent_c}). 
This agent is receiving recommendations on which action to take next, and if a recommendation is provided, then the action is taken.

Figures~\ref{c7:fig:evaluative_persistent_a} and~\ref{c7:fig:evaluative_persistent_b} show evaluative agents, both non-persistent (NPE-*) and persistent (PE-*), using advice from three different users. 
The persistent agent (shown in~\ref{c7:fig:evaluative_persistent_b}) is using PPR to manage the trade-off between the advice received from the user, its own learned policy, and its exploration strategy. 
The persistent agent is limited to only receiving one interaction from the user per state-action pair. 
If the agent has already received some advice for the state-action pair in question, then the user is not given the opportunity to provide additional advice. 
The agent instead relies on the stored advice from the first interaction regardless of its accuracy. 
Both agents will always utilise advice received directly from the user on the current time step. 
However, the persistent agent keeps it and follow a PPR strategy, which allows the agent to diminish the probability of using the advice for a state-action pair over time. 
This results in the persistent agent receiving one interaction per state-action pair, maximising the utility that interaction, then eventually only relying on its own policy. 

The agents being assisted by optimistically initialised users perform almost the same. 
The optimistically-assisted persistent agent (PE-O) takes slightly longer to learn than the non-persistent counterpart (NPE-O), because the advice it receives is only listened to 80\% (diminishing over time) after the initial interaction with the user, compared to the non-persistent agent whose user will continually interact with the agent and the agent will always follow the advice.
The agents being assisted by realistically initialised users differ greatly in performance. 
The non-persistent evaluative agent using a realistically initialised user (NPE-R), while able to solve the mountain car problem in fewer steps than the 1000 cut-off limit, was not able to find the optimal solution. 
However, the persistent evaluative agent (PE-R) was not only able to solve the problem, but also learned the solution faster than the benchmark unassisted agent (UQL), just like the NPE-O and PE-O agents. 
The difference in performance is not only due to the persistent agent remembering the advice, but also because it can eventually disregard incorrect advice as the likelihood that the PPR algorithm will choose to take the recommended action diminishes over time, while the agent's value estimation of the recommended action remains the same. 
What is particularly notable from these results is that the persistent agent (PE-R), still outperformed unassisted Q-learning despite more than half of all interactions giving the incorrect advice.
Regardless of whether the agent is persistent or not, neither agent that was advised by a pessimistically-initialised user (NPE-P, PE-P) was able to solve the mountain car problem. 
This is likely due to the accuracy of the pessimistic user being less than 25\%. 

Figures~\ref{c7:fig:informative_persistent_c} and~\ref{c7:fig:informative_persistent_d} show the performance of informative agents, both non-persistent (NPI-*) and persistent (PI-*), using advice from three users with different levels of advice accuracy and availability. 
These agents can receive informative advice from a user. 
The advice that they receive is an action recommendation, informing them of which action to take in the current state. 
When either agent, persistent or non-persistent, receives an action recommendation directly from the user on the current time step that action will be taken by the agent. 
The persistent agent will remember that action for the state it was received in, and use the PPR algorithm to continue to take that action in the future. 
Once the persistent agent has received an action recommendation from the user for a particular state, the user will not interact with the agent for that state in the future.

\begin{table*}
\caption{Average number of interactions performed per experiment, and the percentage of interactions compared to the total steps taken, for each non-persistent and persistent agent/user combination. }
\begin{center}
\begin{tabular}{|l|l|l|}
\hline
 & \multicolumn{2}{|c|}{\textbf{Interaction}} \\
 & \multicolumn{2}{|c|}{\textbf{(\% Interactions / Total Steps)}} \\
 \hline
 \textbf{Agent} & \textbf{Non-persistent} & \textbf{Persistent}\\
 \hline
 Evaluative agent / Optimistic user (NPE-O/PE-O)  	& 58,355 (100.00\%) & 668 (0.91\%)\\
 Evaluative agent / Realistic user (NPE-R/PE-R)  	& 486,503 (26.86\%) & 117 (0.01\%)\\
 Evaluative agent / Pessimistic user (NPE-P/PE-P)  	& 500,499 (13.43\%) & 47 ($<$0.01\%)\\
 \hline
 Informative agent / Optimistic user (NPI-O/PI-O)  	& 54,083 (100.00\%) & 253 (0.46\%)\\
 Informative agent / Realistic user (NPI-R/PI-R)  	& 134,590 (47.31\%) & 255 (0.01\%)\\
 Infomative agent / Pessimistic user (NPI-P/PI-P)  	& 193,170 (23.65\%) & 63 (0.38\%)\\
 \hline
\end{tabular}
\end{center}
\label{c7:tab:interactions_persistent}
\end{table*}

The informative agents (NPI-O, PI-O), regardless of persistence, learned the solution in the same amount of time when being advised by an optimistically initialised user. 
This is not surprising as the agent is receiving 100\% accurate advice for every time step, making this essentially a supervised learning task at a great effort of the user. 
A difference in the time required to find a solution can be seen in the agents that are assisted by a realistically initialised agent (NPI-R, PI-R). 
While the non-persistent agent (NPI-R) agent does learn faster than an unassisted agent (UQL), the persistent agent learns the solution almost immediately, much like the optimistically-assisted persistent agent (PI-O). 
This difference in learning speed is likely due to the agent retaining and reusing advice. 
The NPI-R and PI-R agents are being assisted by a simulated user with realistic values for accuracy and availability. 
The realistic simulated user has a $\thicksim$48\% chance of interacting with an agent on any particular time step. 
The non-persistent agent does not retain advice from the user, so it will always have a $\thicksim$48\% chance of receiving advice for any particular state. 
However, the persistent agent will retain and reuse advice with an 80\% (diminishing over time from PPR) probability for any state that it has received advice on in the past. 
As long as the retained advice is sufficiently accurate, the persistent agent will learn faster because it utilises that advice more often. 
The last two agents are assisted by a pessimistically initialised user. 
The non-persistent agent outperformed the persistent agent in this experiment. 
This is due to the same principle as the realistically-assisted informative agents. 
The pessimistically-assisted agent performed the recommended action more often than the non-persistent agent. 
Both agents have a 23.6\% chance of receiving advice from the pessimistic user, however, the persistent agent retains and reuses this advice, and will take the recommended action 80\% of the time for states it has been advised on. 
This results in the PI-R agent taking the advised incorrect action far more often than the NPI-R agent.


Table~\ref{c7:tab:interactions_persistent} shows the number and percentage of interactions that occurred on average for each agent/user combination. 
However, the number of interactions is not suitable to compare agents, as agents that benefit from advice may take fewer steps, giving the users fewer opportunities to provide advice, despite perhaps requiring more attention from the user per episode. 
Therefore, the percentage of interaction is more suitable for comparing agents against each other, as it is a function of the interaction requirements, rather than a direct measurement of the number of interactions. 
For non-persistent agents, this interaction percentage is equal to the advising user availability. 
For persistent agents, this percentage varies due to the use of the PPR approach.

It is clear from Table~\ref{c7:tab:interactions_persistent} that persistent agents require substantially fewer interactions than non-persistent agents. 
These results show that the number of interactions required by the user to achieve each agent's recorded performance is significantly reduced when advice is retained.  
All persistent agents measured less than 1\% of steps with direct user interaction. 
Assuming a direct correlation between the number of interactions and the time required to perform those interactions, the use of persistence offers a large time reduction for assisting users. 
This significant drop in required interactions, coupled with the previous observation of large performance gains shown by the majority of persistent agents, makes a compelling case for the retention and reuse of advice, assuming a suitable level of accuracy of that advice.

For non-persistent agents, an observation can be made that as the availability of the simulated user decreases, the number of interactions increases. 
%
In this case, simulated users that are highly accurate allow the agent to learn the optimal policy faster, which results in the agent taking fewer steps, and the simulated user having fewer opportunities to interact. 
Simulated users with lower accuracy, such as the pessimistic users, cause the agent to take longer to learn the policy, resulting in the agent taking more steps, and allowing the simulated user more opportunities to provide advice. 
This is what creates the inverse correlation between the user advice availability and the number of interactions recorded in non-persistent agents.
The same situation is not observed for persistent agents. 
This is due to the use of the PPR approach, which leads to similar opportunities to provide advice in all the cases. 
For instance, if advice is given in a state that has previously received, this might be dismissed by the PPR approach delimiting the total number of interactions regardless of the accuracy of such advice.

\begin{figure*}[!hbt]
\centering
\subfloat[State-based agent.]{\includegraphics[width=0.5\linewidth]{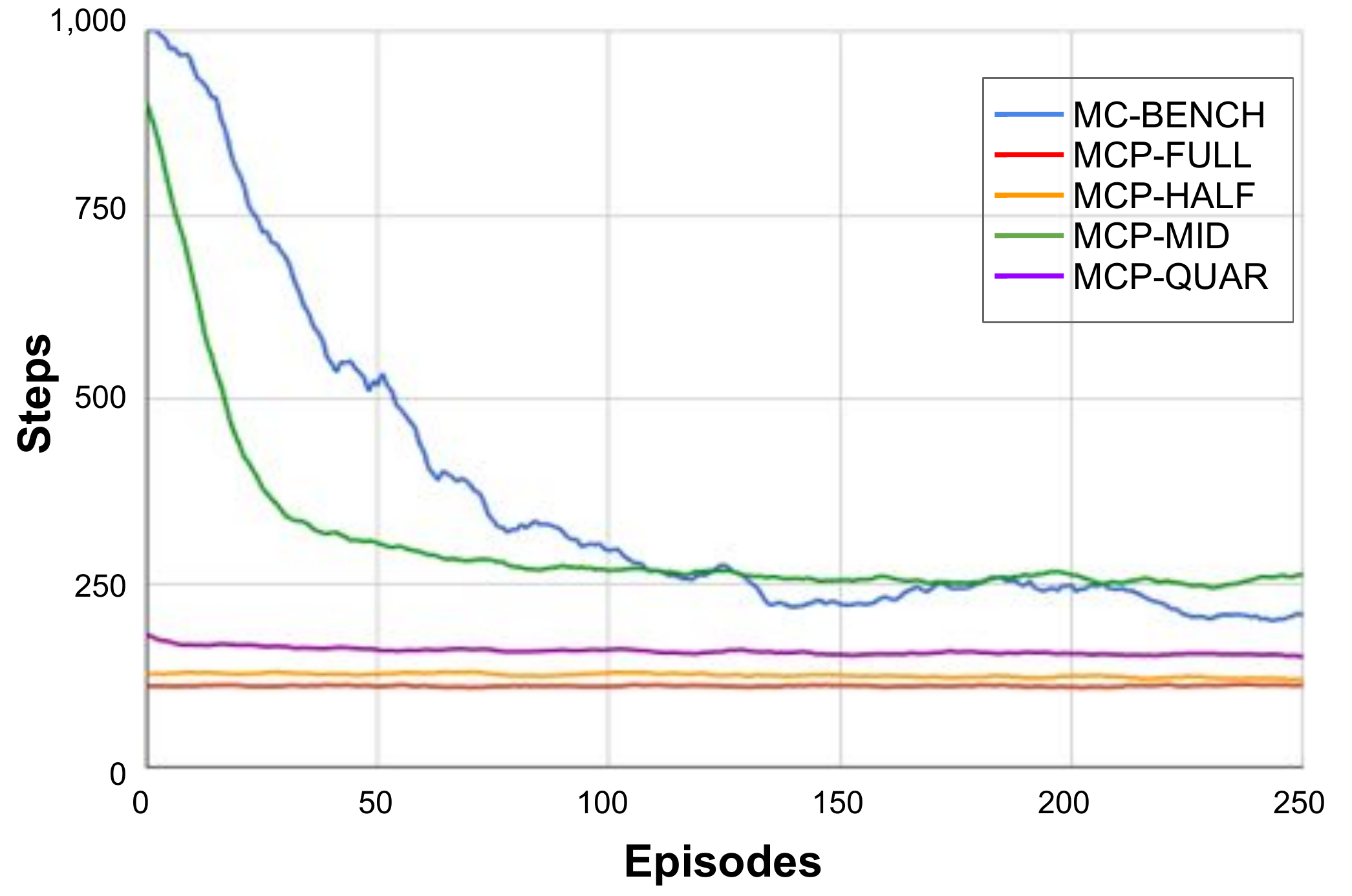}
\label{fig:mc_rdr_steps_a}} 
\subfloat[Rule-based agent.]{\includegraphics[width=0.5\linewidth]{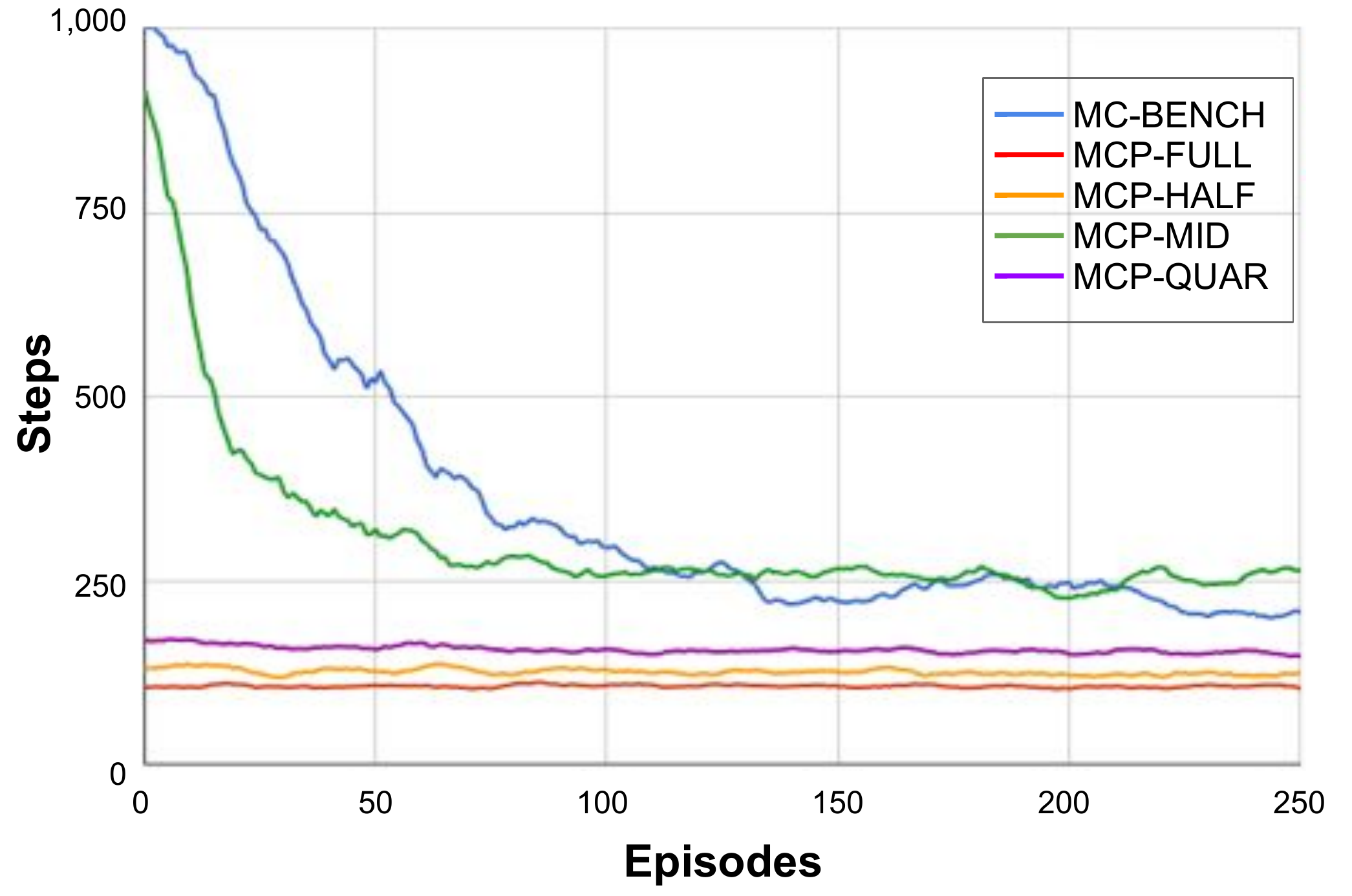}
\label{fig:mc_rdr_steps_b}}
\caption{Step performance for state-based and rule-based IntRL agents for the mountain car domain. Using considerable much less interaction from the trainer, results show no significant difference in performance between the two types of agents.}
\label{fig:mc_rdr_steps}
\end{figure*}

\subsection{Rule-based Advice}
Using a rule-based persistent advice technique, we expect to reduce even further the number of interactions needed between the learning agent and the trainer, in comparison to the state-based persistent advice.
In this regard, we perform experiments using two different domains, namely, the mountain car domain and the self-driving car domain described in Section 4.

\subsubsection{Mountain Car Domain}
First, we employ the mountain car domain using both state-based and rule-based advice.
A total of eight different simulated users are created, four are used for state-based advice and four for rule-based advice.
The agents differ in the level of knowledge and the availability to deliver advice, i.e., full, half, quarter, and bottom availability (as shown in Table \ref{c8:tab:sim_user_settings_rule}).
Figure~\ref{fig:mc_rdr_steps} shows the number of steps each agent performed each episode for this environment. 
Figure~\ref{fig:mc_rdr_steps_a} shows the results for the state-based agents, and Figure~\ref{fig:mc_rdr_steps_b} shows the rule-based agents. 
A comparison of the two graphs shows that the agents performed similar, regardless of the advice delivery method. 
This was expected, as the method in which the agent uses the advice and the amount of advice in total that the agent receives does not differ between the two types of agents.  
The agents using minimal advice (MCP-MID and MCRDR-MID) end up learning a worse behaviour than the unassisted Q-Learning agent. 
This is likely an indication that the decay rate for the PPR action selection method is too low, and that the agent has not yet learned to ignore the user advice after its initial benefit and focus on its own learning.

Table~\ref{c8:tab:mc_rdr_interactions} shows the number of interactions, and the percentage of interactions over opportunities for interactions, for each agent. 
These results show that the number of interactions is much less for the rule-based agents compared to the state-based agents, allowing similar performance with much less effort from the trainer. 
In the previous experiment, the number of interactions was not a useful measure to compare agents against each other. 
This was because the advice provided to the agent affects the number of steps the agent takes, which results in fewer opportunities for interactions. 
However, Figure~\ref{fig:mc_rdr_steps} shows that the performance the agents that use the same simulated user are the same, regardless of the advice type. 
Therefore, in this context, the number of interactions is a useful measure for comparing the corresponding state-based and rule-based agents.

\begin{table}
\caption{Interaction percentage state- and rule-based agents for the mountain car domain. Average number of interactions performed per experiment, and the percentage of interactions compared to the steps taken, for each state-based/rule-based agent/user combination.}
\begin{center}
\begin{tabular}{ |p{5.6cm}|p{2.2cm}|  }
 \hline
 \textbf{Agent / User} & \textbf{\#Interaction (\%)}\\
 \hline
 State-based / Full (MCP-FULL)		&254 ($<$0.01\%)\\
 State-based / Half (MCP-HALF)		&227 ($<$0.01\%)\\
 State-based / Quarter (MCP-QUAR)	&139 ($<$0.01\%)\\
 State-based / Bottom (MCP-MID)		&45 ($<$0.01\%)\\
 \hline
 Rule-based / Full (MCRDR-FULL)	&2 ($<$0.01\%)\\
 Rule-based / Half (MCRDR-HALF)	&3 ($<$0.01\%)\\
 Rule-based / Quarter (MCRDR-QUAR)	&3 ($<$0.01\%)\\
 Rule-based / Bottom (MCRDR-MID)		&3 ($<$0.01\%)\\
 \hline
\end{tabular}
\end{center}
\label{c8:tab:mc_rdr_interactions}
\end{table}

\begin{figure}[!hbt]
\centering
\subfloat[Steps per episode.]{\includegraphics[width=1\linewidth]{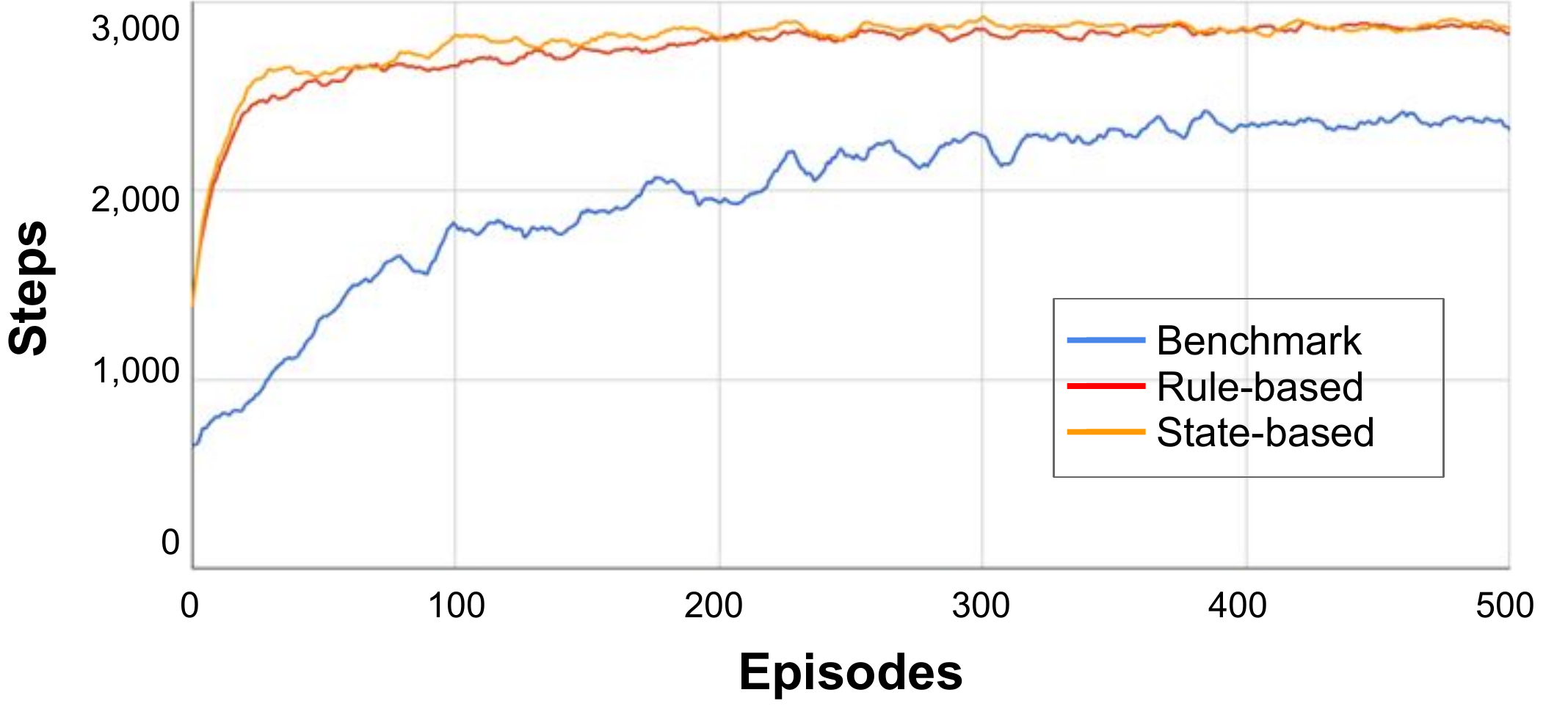}
\label{fig:sdc_rdr_steps_a}} 
\\
\subfloat[Reward per episode.]{\includegraphics[width=1\linewidth]{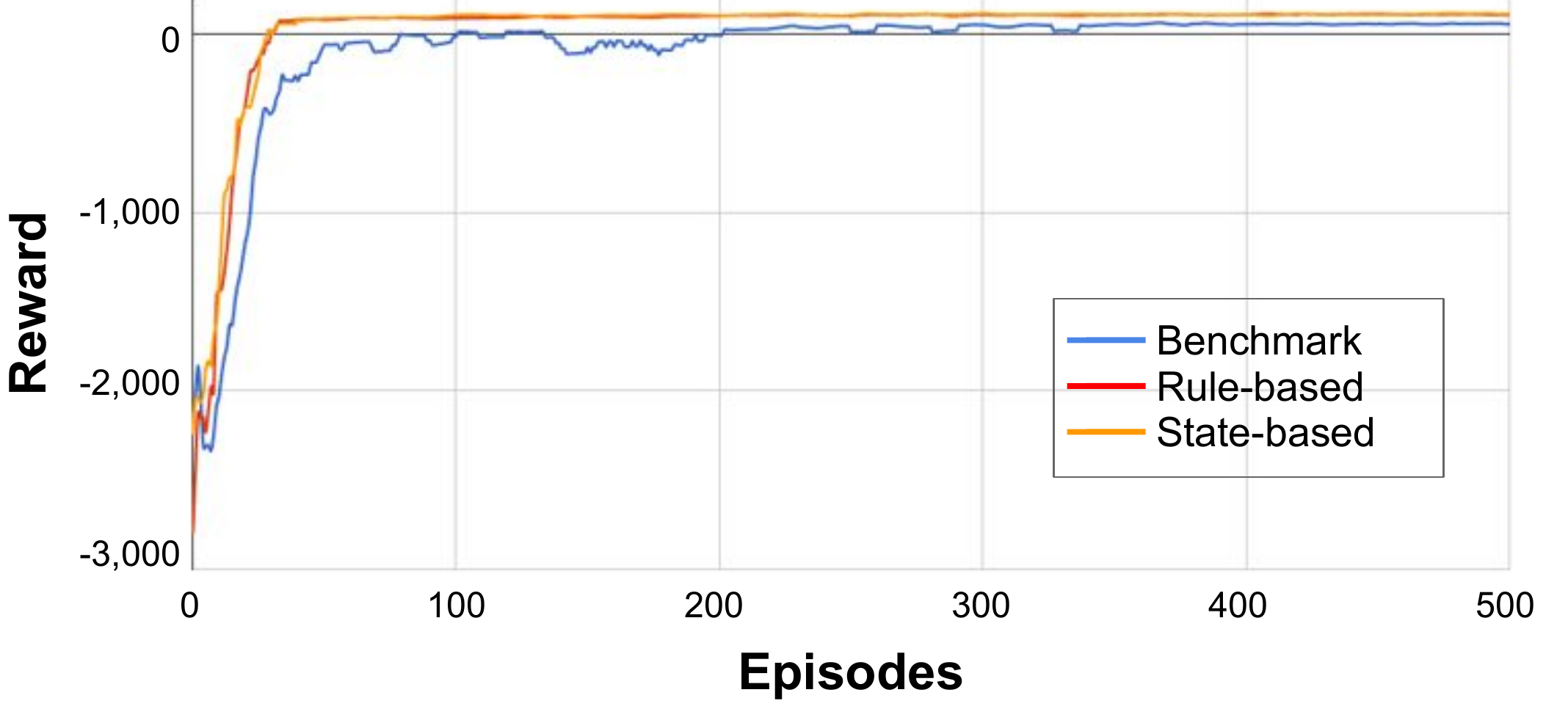}
\label{fig:sdc_rdr_reward_b}}
\caption{Steps and reward for state-based and rule-based IntRL agents for the self-driving car domain. The advice required from the trainer is considerable less obtaining no significant difference in performance between the two types of agents.} 
\label{fig:sdc_rdr}
\end{figure}

\subsubsection{Self-driving Car Domain}

The aim of the agent in the self-driving car environment is to avoid collisions and maximise speed. 
In the experiments, we created two simulated users to provide state-based and rule-based advice. 
Both agents outperformed the unassisted Q-Learning agent, both achieving a higher step count and reward. 
The obtained steps and reward are shown in Figure~\ref{fig:sdc_rdr_steps_a} and Figure~\ref{fig:sdc_rdr_reward_b} respectively.
Although the agent was forcibly terminated when it reached 3000 steps, Figure~\ref{fig:sdc_rdr_steps_a} shows that the agents never reached the 3000 step limit.
This is because the agents are given a random starting position and velocity at the beginning of each episode, some of which result in scenarios where the agent cannot avoid a crash.
Although both assisted agents outperformed the unassisted agent, between the state-based and rule-based methods, there is no considerable difference since both run a similar number of steps and collected a similar reward.

Table~\ref{c8:tab:sdc_rdr_interactions} shows the number of interactions, and the percentage of interactions compared to the number of opportunities for interactions (equal to steps), for each agent. 
These results show that the number of interactions is much less for the rule-based agents compared to the state-based agents. 

\begin{table}
\caption{Average number of interactions performed per experiment and the percentage of interactions compared to the steps taken, for each state-based/rule-based agent/user combination in the self-driving car domain.}
\begin{center}
\begin{tabular}{ |p{4.7cm}|p{2.5cm}|  }
 \hline
 \textbf{Agent} & \textbf{\#Interaction (\%)}\\
 \hline
 State-based advice agent	&232 ($<$0.01\%)\\
 \hline
 Rule-based advice agent 	&2 ($<$0.01\%)\\
 \hline
\end{tabular}
\end{center}
\label{c8:tab:sdc_rdr_interactions}
\end{table}

\section{Conclusions}
In this work, we have introduced the concept of persistence in interactive reinforcement learning. 
Current methods do not allow the agent to retain the advice provided by assisting users. 
This may be due to the effect that incorrect advice has on an agent's performance. 
To mitigate the risk that inaccurate information has on agent learning, probabilistic policy reuse was employed to manage the trade-off between following the advice policy, the learned policy, and an exploration policy. 
Probabilistic policy reuse can reduce the impact that inaccurate advice has on agent learning.

Interactive reinforcement learning agents, both evaluative and informative, learned faster when retaining the information provided by an advising user, when the advising user's accuracy is sufficient. 
Additionally, persistent agents were shown to require significantly fewer interactions than non-persistent agents, while achieving the same or better learning speeds when advice accuracy was sufficient. 

Additionally, we have introduced rule-based interactive reinforcement learning, a method for users to assist agents through the use of rule-structured advice and retention. 
Two environments were tested to investigate the impact that rule-based advice had on performance and the number of interactions performed to achieve the measured performance. 
Compared to state-based persistent advice for interactive reinforcement learning, rule-based advice was able to achieve the same level of performance with substantially fewer interactions between the agent and the user. 

This work did not investigate the time and cognitive requirements for users to construct state-based and rule-based advice. 
It is likely that rule-based advice will require more time and thought to construct. 
However, existing research has shown that decision trees built with ripple-down rules are easier for users to construct~\cite{gaines1995induction, compton1991ripple, compton2006experience}. 
Future work is required to test if this will justify the benefits that rules provide over state-based advice, in terms of the number of interactions.

\section*{Acknowledgments}
This work has been partially supported by the Australian Government Research Training Program (RTP) and the RTP Fee-Offset Scholarship through Federation University Australia.

\bibliographystyle{ieeetr}
\balance
\bibliography{paper}

\end{document}